\def\paperTitle{Analyzing Local Representations of Self-supervised Vision Transformers}

\def\authorBlock{
    Ani Vanyan  \thanks{YerevaNN, Yerevan State University (YSU)} \qquad
    Alvard Barseghyan \footnotemark[1] \thanks{CSIE}\qquad
    Hakob Tamazyan \footnotemark[1] \qquad \\
    Vahan Huroyan \footnotemark[1] \qquad 
    Hrant Khachatrian \footnotemark[1] \qquad 
    Martin Danelljan \thanks{ETH Zurich} \\
    {\tt\small \{ani, alla, hakob, vahan, hrant\}@yerevann.com}
    
}

\newif\ifreview 
\newif\ifarxiv \newcommand{\arxiv}{\arxivtrue}
\newif\ifcamera 
\newif\ifrebuttal 

\arxiv
\pdfoutput=1
\documentclass[10pt,letterpaper]{article}
\ifreview \usepackage[review]{cvpr} \fi
\ifarxiv \usepackage[pagenumbers]{cvpr} \fi
\ifrebuttal \usepackage[rebuttal]{cvpr} \fi
\ifcamera \usepackage{cvpr} \fi

\usepackage{graphicx}	
\usepackage{amsmath}	
\usepackage{amssymb}	
\usepackage{booktabs}
\usepackage{times}
\usepackage{microtype}
\usepackage{epsfig}
\usepackage[table,xcdraw,dvipsnames]{xcolor}
\usepackage{caption}
\usepackage{float}
\usepackage{placeins}
\usepackage{color, colortbl}
\usepackage{stfloats}
\usepackage{enumitem}
\usepackage{tabularx}
\usepackage{xstring}
\usepackage{multirow}
\usepackage{xspace}
\usepackage{url}
\usepackage{subcaption}
\usepackage{xcolor}
\usepackage[hang,flushmargin]{footmisc}

\ifcamera \usepackage[accsupp]{axessibility} \fi

% ...

\ifarxiv  \fi

\newcommand{\R}[1]{{%
    \textbf{%
        \ifstrequal{#1}{1}{\textcolor{red}{R#1}}{%
        \ifstrequal{#1}{2}{\textcolor{blue}{R#1}}{%
        \ifstrequal{#1}{3}{\textcolor{magenta}{R#1}}{%
        \ifstrequal{#1}{4}{\textcolor{teal}{R#1}}{%
                           \textcolor{cyan}{R#1}%
        }}}}%
    }%
}}
\usepackage{xr-hyper}

\makeatletter
\newcommand*{\addFileDependency}[1]{
  \typeout{(#1)}
  \@addtofilelist{#1}
  \IfFileExists{#1}{}{\typeout{No file #1.}}
}

\makeatother
\newcommand*{\myexternaldocument}[1]{
    \externaldocument{#1}
    \addFileDependency{#1.tex}
    \addFileDependency{#1.aux}
}

\definecolor{cvprblue}{rgb}{0.21,0.49,0.74}
\usepackage[pagebackref,breaklinks,colorlinks,citecolor=cvprblue]{hyperref}
\usepackage[capitalize]{cleveref}
\crefname{section}{Sec.}{Secs.}
\crefname{table}{Table}{Tables}
\crefname{figure}{Fig.}{Figs.}

\ifarxiv \crefname{appendix}{App.}{Apps.}
\else \crefname{appendix}{Suppl.}{Suppls.} \fi

\unless\ifarxiv \myexternaldocument{_supplementary} \fi

\begin{document}

% ---------------------------------------------------------------

%% TITLE
\title{\paperTitle}
\author{\authorBlock}
\maketitle

\begin{abstract}
  In this paper, we present a comparative analysis of various self-supervised Vision Transformers (ViTs), focusing on their local representative power. Inspired by large language models, we examine the abilities of ViTs to perform various computer vision tasks with little to no fine-tuning.
  We design evaluation framework to analyze the quality of local, i.e.\ patch-level, representations in the context of few-shot semantic segmentation, instance identification, object retrieval and tracking. 
  We discover that contrastive learning based methods like DINO produce more universal patch representations that can be immediately applied for downstream tasks with no parameter tuning, compared to masked image modeling. The embeddings learned using the latter approach, e.g. in masked autoencoders, have high variance features that harm distance-based algorithms, such as k-NN, and do not contain useful information for most downstream tasks. 
  Furthermore, we demonstrate that removing these high-variance features enhances k-NN for MAE, as well as for its recent extension Scale-MAE.
  %by providing an analysis of the benchmarks for this work and for Scale-MAE, a recent extension of masked autoencoders.
  Finally, we find an object instance retrieval setting where DINOv2, a model pretrained on two orders of magnitude more data, 
  % performs worse than its less compute intensive counterpart DINO.
  falls short of its less compute intensive counterpart DINO.
  % \keywords{SSL \and ViT \and Representation Learning}
\end{abstract}

\section{Introduction}

Recent advances in Natural Language Processing gave birth to universal models that after large-scale pretraining can perform various language-related tasks without task-specific fine-tuning. Large language models \cite{GPT3,GPT4, unreasonable-machine-translation} based on self-supervised transformers achieve competitive performance on tasks like translation, question answering, and commonsense reasoning with prompting or by in-context learning with just a few examples. 
Self-supervised transformers are also getting increasingly popular in computer vision. Two radically different self-supervised learning paradigms have demonstrated good performance for Vision Transformers (ViTs): those based on contrastive learning (e.g. MOCO~\cite{MOCO} or DINO~\cite{DINO}),
% , such as DINO~\cite{DINO}
and those based on masked image modeling (e.g. MAE~\cite{MAE} or SimMIM~\cite{SimMIM}).
The question of whether these models possess universal capabilities, similar to those seen in NLP models, for computer vision remains unanswered. 

As ViTs do not have text inputs, it is non-trivial to assess their zero-shot capabilities for downstream tasks. Most ViTs produce one embedding vector for the entire image (usually the [CLS] token) and one embedding for each local patch. %While these patch representations correspond to specific locations inside the image, they contain some global context as well. 
In this paper we focus on few-shot capabilities of ViTs for vision tasks that require locality awareness, like image segmentation and object tracking. We propose few-shot evaluation methods that leverage patch representations. To minimize task-specific parameter tuning we use two approaches: k-NN and linear probing with a single layer of trainable parameters. The power of global image representations from pretrained ViTs for image-level tasks like image classification are relatively well explored in literature, e.g. \cite{DINO}. 

Fine-tuning the entire backbone in addition to relatively large task-specific ``heads'' still gives superior performance in segmentation and tracking. The analysis of such models is beyond the scope of this paper as their good performance is not just from self-supervised pretraining, but also strongly affected by the head architecture and the data used for fine-tuning. The focus of this work is on the inherent capabilities of self-supervised ViTs that can be exposed by using just a few labeled samples.
We show that while masked image modeling produces backbones with good fine-tuning performance, the frozen, pretrained patch embeddings are far inferior to the ones learned by contrastive methods for nearest neighbor methods. We dive deep into this phenomenon and identify roughly 200 dataset-agnostic features in the embedding space that, counterintuitely, contain no useful information for the downstream tasks we have considered, while having the highest variances among all features. Removing those features improves k-NN performance for most tasks. 

% Finally we examine the effect of pretraining ViTs on larger datasets. While we see improvements in many aspects, including robustness with respect to degradations, we also identify settings when the model trained on more data performs worse than its less  

We further explore the usefulness of patch embeddings for identifying the same object instance in multiple images. We perform experiments on a satellite imagery dataset under several image transformations, and find that DINO surprisingly outperforms its newer and larger sibling DINOv2~\cite{DINOv2}. Additionally, we measure the quality of patch embeddings in distinguishing fine-grained object categories. Lastly, we perform experiments for object association on multi-object tracking datasets. We find that DINO and DINOv2 substantially outperform both masked image models and supervised ViTs, making them most suitable for object retrieval in videos. Our main contributions are summarized as follows:

\begin{itemize}
    \item We design an evaluation framework along with few-shot datasets to analyze inherent power of pretrained ViTs for locality aware tasks. We analyze and compare five ViTs using our framework on three tasks: patch classification, instance/fine-grained retrieval, and object association in video frames.
    \item We show that compared to masked image modeling, contrastive pretraining produces significantly more universal patch embeddings that can be immediately utilized in downstream tasks without fine-tuning. We identify the cause of poor performance of MAE-like models in methods based on k-NN. 
    Upon removing the high-variance features (200 in our experiments), the performance of MAE-like models for k-NN shows a significant improvement.
    \item 
    We demonstrate that the removal of these features is beneficial not only for our proposed benchmarks but also for benchmarks presented in other studies. For instance, Scale-MAE~\cite{scalemae}, which assesses a MAE-like network trained on aerial images, evaluates its performance across various resolutions and compares it against other state-of-the-art algorithms using k-NN on computed representations. We observe the same issue of high-variance features, and after removing 200 such features, we showcase a superior performance.
    \item We show that DINOv2, trained on two orders of magnitude more unlabeled data, outperforms all other ViTs in most scenarios, including in robustness of patch classification with respect to image corruptions. Surprisingly, it underperforms most ViTs in identifying the patches covering the same object instance in transformed images, indicating that blindly adding more data to pretraining might not universally improve all results.
\end{itemize}

\section{Related work}

The recent advent of Visual Transformers (ViT)~\cite{ViT, DINO, DINOv2} and its use in many downstream tasks has paved the way for novel approaches in several directions of computer vision, including 
image segmentation~\cite{Mask2Former}, 
image classification~\cite{ViT}, 
and object detection~\cite{Mask2Former}.
% Action Prediction~\cite{}
% Depth & Flow Estimation~\cite{}.
Unlike Language Models, where the model sizes reached to 175B parameters, scaling ViTs is notoriously difficult~\cite{vit22b}. Both works DINOv2~\cite{DINOv2} and ViT-22B~\cite{vit22b} claim that their core technical contribution is about stabilizing the training of large transformers on hundreds of millions of images.

\cite{park2022how} analyze and demonstrate several properties of multi-head self-attentions (MSA) and ViTs. They show that MSAs flatten the loss landscape to alleviate the issue of its non-convexity.
% It is known that ViTs suffer from non-convex losses, but large datasets and loss landscape smoothing methods help alleviate this issue.
They also observe that MSAs and convolutional layers complement each other, showing that MSAs function as low-pass filters, while convolutional layers act as high-pass filters.
~\cite{what-SS-ViT-learns-iclr2023} analyze the differences between the ViT methods that are based on contrastive learning (CL) and masked image modeling (MIM) and compare their performance of downstream tasks. 
They demonstrate that CL captures longer-range global patterns, such as object shapes, more effectively than MIM methods. 
Secondly, they demonstrate that CL-based approaches are more shape-oriented, whereas MIM-based approaches are more texture-oriented.
\cite{vits_see_like_cnn} provide an analysis of the internal representation structure of ViTs and CNNs on several image classification benchmarks. 
They demonstrate that ViTs have a more uniform representation across the layers of the network compared to CNNs. These differences are mostly explained by the role of self-attention, which allows early aggregation of information, and ViT residual connections, which also propagate features from lower to higher levels.

Other works have focused on analyzing the robustness of ViTs.
~\cite{bhojanapalli2021understanding} investigate the robustness of ViT models to input and model perturbations for image classification. ~\cite{bhojanapalli2021understanding} demonstrate that transformers are robust to the removal of almost any single layer and that when pre-trained on a sufficiently large dataset, ViTs demonstrate not worse results than the ResNet counterparts across various perturbations.
~\cite{paul2022vision} analyze the robustness of ViTs against several common corruptions, perturbations, distribution shifts, and natural adversarial examples. They also analyze and demonstrate the superior robustness of ViTs in various aspects, such as masking, energy/loss landscape analysis, and sensitivity to high-frequency artifacts on robust classification datasets.
%Naseer et al.~
\cite{naseer2021intriguing} investigate the robustness of Transformers to severe occlusions, perturbations, and domain shifts in classification tasks. Their findings demonstrate that ViTs exhibit significantly less bias towards local textures compared to CNNs.

Another line of research, that contributes to the universality of the models and enables zero-shot image classification (and potentially other vision tasks) involves vision-language models, including contrastive models like CLIP~\cite{CLIP} and autoregressive models like CM3Leon~\cite{CM3leon}. Analysis of these models is beyond the scope of this paper.

While not precisely the same phenomenon, a similar occurrence of artifacts in feature maps is analyzed in \cite{registers}. The study demonstrates that for supervised and self-supervised Vision Transformers (VITs), certain artifact high-norm tokens emerge during inference, primarily in low-informative background areas of images. It is worth noting that our observation for MAE is the presence of features with high variance, whereas \cite{registers} observes that certain tokens exhibit artifacts.
% \begin{wrapfigure}{r}{0.45\textwidth}
\begin{figure}
    \centering
    % \vspace{-5ex}
    \includegraphics[width=0.4\textwidth]{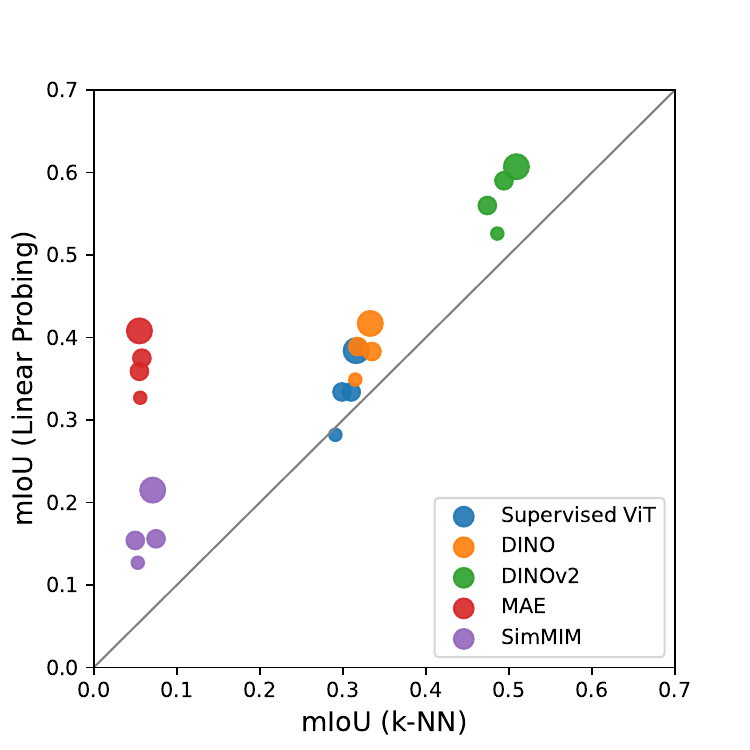}
    \caption{mIoU of patch classifiers on various subsets of Cityscapes dataset. The evaluation is performed on the same validation set. The size of the circles are proportional to the number of images in the training set (36, 72, and 144).}
    % \vspace{-10ex}
    \label{fig:training-set-size}
\end{figure}
% \end{wrapfigure}
Our work analyzes and compares different ViTs regarding their ability to locally represent images. We explore and compare if the local patch representations obtained from ViTs trained with different self-supervised or supervised strategies. To this end, we probe the quality of the patch-wise features for dense patch classification, fine-grained retrieval, and tracking, in a few-shot setting.

\section{Evaluation Frameworks}

Throughout the paper we use five ViT models. MAE \cite{MAE} and SimMIM~\cite{SimMIM} are used as representative models of masked image modeling. Contrastive models are represented by DINO~\cite{DINO} and its counterpart DINOv2~\cite{DINOv2} which unlike all other ViTs used in this work is pretrained on a much larger dataset than ImageNet. We use Supervised ViT~\cite{ViT} as a baseline, and in one setting we also use iBOT~\cite{ibot} which is trained on ImageNet like DINO but uses loss terms more similar to DINOv2. These models are described in detail in \cref{sec:app-choice-of-vits}.

To fairly assess the effectiveness of existing ViTs, and compare them with our proposed MAE-200 with improved representations \cref{sec:mae_200}, it is important to establish a comprehensive evaluation framework. 
The objective of our proposed evaluation framework is assessing the effectiveness of existing SSL methods for several downstream tasks, using linear probing and k-NN.
We consider 3 different setups for evaluations. 
The first setup includes analyzing the local representations of the ViTs, by performing patch-wise classification. 
To this end, we set up a few-shot patch classification experiment on the Cityscapes dataset~\cite{cityscapes}; see Sec.~\ref{sec:results}.
The second setup analyzes the performance of ViTs for object-level few-shot semantic segmentation task. 
This task involves classifying patches with predefined objects like airplanes and cars. We explore whether ViTs can distinguish between different types within the same category or effectively differentiate between instances of the same object with multiple occurrences. We use the FAIR1M dataset~\cite{sun2021fair1m}; see \cref{sec:results}.
% This task entails assigning a class to each patch containing predefined objects, such as airplanes and cars. We further investigate whether ViTs can learn to distinguish between objects of the same category but different types (such as different types of cars or airplanes), or whether ViTs can effectively differentiate between instances of the same object when multiple instances are present.  We utilize the FAIR1M dataset~\cite{sun2021fair1m}; see Sec.~\ref{sec:results}.
In the third setup, we assess the ability of ViTs to track objects. Object tracking involves identifying the same object instance across frames in a video. We analyze the robustness of patch embeddings over time, especially as objects undergo appearance changes. 
We utilize the track-validation set of MOT 2020 images, a subset of the BDD-100K dataset~\cite{bdd100k}; see Sec.~\ref{sec:results}.
At last, in Sec.~\ref{sec:discussion}, we discuss additional numerical experiments under different setups to support our claims.

% \noindent\textbf{Few-shot subset of Cityscapes.}
% \subsection{Cityscapes Dataset}
% \label{sec:cityscapes}

\noindent\textbf{Cityscapes Dataset~\cite{cityscapes}.}
% To analyze the local representations of the ViT models, we first study their ability to perform patch-wise classification. To this end, we set up a few-shot patch classification experiment on the Cityscapes dataset~\cite{cityscapes}. 
The training set contains 2975 pictures taken in 18 cities. Unless otherwise stated, we use a training dataset consisting of 4 images per city (72 images in total). 
% We also explore the impact of the number of training samples by varying this number.
The original validation set has $500$ images, containing images from 3 different cities. 
% For our analyses, w
We take $10$ images per city, resulting in a total of $30$ images.
We convert the pixel-dense segmentation labels to patch-level classes by selecting the most common class within each patch.
We evaluate the quality of representations by measuring pixel accuracy and segmentation mIoU. 
As the ViTs that we used take as inputs images of size $224\times224$, we tile the images of size $1024 \times 2048$ into $256\times256$ tiles and treat each tile as a separate image. 
The tiles are further resized to $224\times224$.
% and passed to the pre-trained transformers.
% We extract and store the patch representations of the corresponding transformer for all the patches in all the images (training and validation) and for all the ViTs.
We store patch representations for all images (training and validation) and ViTs.

% \subsection{Fair1M Dataset}
% \label{sec:fair1m}
\noindent\textbf{Fair1M~\cite{sun2021fair1m}} 
is a satellite imagery dataset designed for fine-grained object detection. 
Note that none of the ViTs we tested were pretrained on satellite imagery (even DINOv2, as far as we can tell).
The objects in FAIR1M are annotated in 5 supercategories: airplane, ship, vehicle, court, and road, and 37 fine-grained categories (types of airplanes, types of ships, etc.). 
The annotations are rotated bounding boxes (without pixel precision).  
To the best of our knowledge, all object instances appear on only one image.
The images have varying sizes, usually larger than $1000 \times 1000$px. We cropped them to $224\times224$px tiles and kept only 8 images per each fine-grained category to 
ensure a minimum of 8 instances for each category.
% guarantee that our dataset contains at least 8 instances of each category, but in fact it contains many more instances of the common objects. 
\cref{tab:fair1m-stats} in Appendix lists dataset statistics. 
We have 295 images with 196 patches each for ViT-B/16 models, and 256 patches each for DINOv2.

% \subsection{Tracking Dataset}
\noindent\textbf{Tracking Dataset.}
For the tracking experiments, we take track-validation set of MOT 2020 images subset of BDD-100K dataset~\cite{bdd100k} and extract patch representations from each frame. We use ROIAlign~\cite{mask-rcnn} to pool features of each object instance based on its ground truth bounding box. For a given time difference $\Delta$ between two frames and for each frame at time step $t$, we retrieve the object with the closest embedding vector from the frame at time step $t + \Delta$.

\section{k-NN Fails for MIMs: Proposed Solution}
\label{sec:mae_200}

In this section, we tackle a challenge with MIM-based methods in k-NN tasks. 
The problem revolves around features that have significantly higher variance, compared to other features. 
We suggest a solution for this issue and then, in~\cref{sec:discussion}, we take a closer look at what information these features hold.

% \noindent\textbf{k-NN and linear probing.}
\noindent\textbf{Issue Discovery: k-NN vs. Linear Probing.}
To introduce the issue that we have noticed, we conduct the  following experiment: 
We use two simple classifiers to analyze the underlying representations: k-NN with $k = 1$ and fitting a linear softmax classifier. 
Both are trained on the patch representations of the few-shot training set. 
The motivation for employing these two basic methods is to understand whether the patches of a given object category cluster together or are linearly separable from other object categories in the representation space.
For MAE, we follow recommendations from~\cite{MAE} (which on its turn refers to \cite{context_predictor}) and apply batch normalization to the extracted features before the linear layer achieving $2.5\times$ better results for linear probing with BatchNorm. We do not use BatchNorm for SimMIM, as adding it worsens the performance.
The results are demonstrated in Fig.~\ref{fig:training-set-size}.
We observe that MAE and SimMIM perform significantly worse for k-NN compared to linear probing. Since both methods employ pixel-level reconstruction loss, we hypothesized that something may be wrong with the obtained representations and proceeded to analyze their performance further.

Throughout our experiments we discovered that MAE's patch embeddings perform reasonably well with linear probing, but fail with k-NN.
We hypothesize that the patch embeddings of MAE might have high variance in some dimensions which could drastically increase the distances among same-category patches and therefore harm k-NN, while not affecting the performance of linear models.

\begin{figure*}[t]
    \centering
    \begin{subfigure}[b]{0.30\textwidth}
        \centering
        \includegraphics[width=\textwidth]{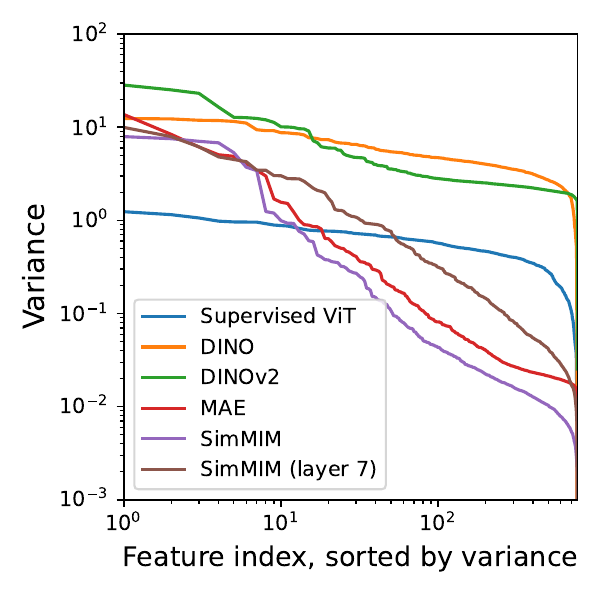}
        \caption{Variances of all (768) features of embeddings of various ViTs sorted in decreasing order. Both axis are logarithmic.}
        \label{fig:variances-all-models}
    \end{subfigure}
    \hfill
    \begin{subfigure}[b]{0.30\textwidth}
        \centering
        \includegraphics[width=\textwidth]{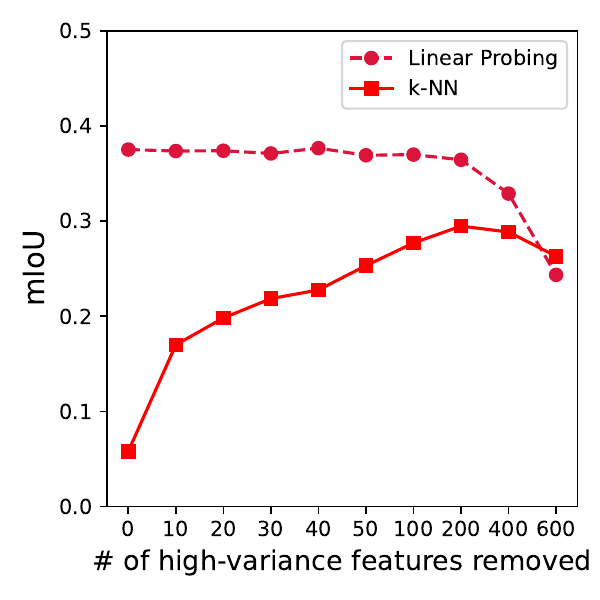}
        \caption{k-NN and linear probing performance of MAE embeddings with $m$ highest-variance dimensions removed.}
        \label{fig:mae-nonhighvariance}
    \end{subfigure}
    \hfill
    \begin{subfigure}[b]{0.30\textwidth}
        \centering
        \includegraphics[width=\textwidth]{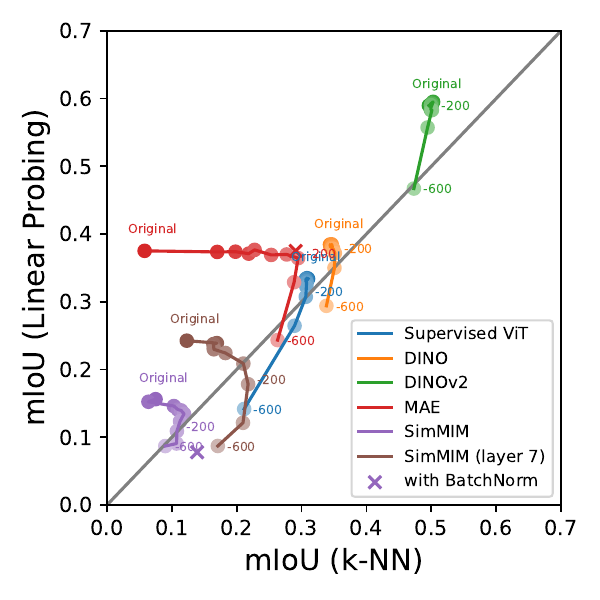}
        \caption{k-NN and linear probing performance of embeddings of various ViTs with $m$ highest-variance dimensions removed.}
        \label{fig:nonhighvariance}
    \end{subfigure}
    \caption{Analysis of high-variance features of ViT patch embeddings.}
    \label{fig:variance}
\end{figure*}

% \subsection{Why MIM-based models have poor k-NN performance?}
\noindent\textbf{Why MIM-based models have poor k-NN performance?} 
% We observe that the variances of MAE and SimMIM features are extremely diverse.
We computed the variance for each of the features for all models. In \cref{fig:variances-all-models}, we plot the variances of all 768 features in a decreasing order. We observe that the variances throughout all features of Supervised ViT (and DINO) are relatively uniform. However, for MAE and SimMIM, there are several features with very high variances and there is a long tail of close to zero variance features.

% \subsection{Simple Remedy}

\noindent\textbf{Simple Remedy:}
\noindent\textbf{Improving k-NN performance for MAE.}
We removed $m$ features with the highest variances and measured few-shot segmentation performance using k-NN and linear probing with the shortened embeddings. As seen in \cref{fig:mae-nonhighvariance}, with just $m=10$ features removed, MAE's k-NN performance jumped from $0.058$ to $0.170$, without sacrificing the performance of linear probing. The performance of k-NN kept increasing up to $0.295$ at $m=200$. Afterwards, the scores for both k-NN and linear probing started to decline. 
This finding implies that around a quarter of features of MAE's embeddings are not useful for patch-level image segmentation for neither linear models nor k-NNs. On the other hand, these features comprise nearly all 
of the variance of the embeddings.

% \subsection{Connections to Dimensional Collapse.}
\noindent\textbf{Connections to Dimensional Collapse.}
The findings mentioned above align with that in~\cite {dimensional_collapse}, which asserts that some contrastive learning algorithms experience dimensional collapse. This occurs when the representation space fails to utilize its entirety, instead maps all images to a lower-dimensional subspace.
Our findings indicate significantly higher variances in some features, suggesting that 
% the space of these features encompasses almost all of the variance. Consequently, 
the representations lie in the space spanned by these high-variance features. 
We show that by removing these features, the performance of MAE for downstream tasks improves for k-NN and does not deteriorate for linear probing. 
These findings are somewhat orthogonal to that of~\cite {dimensional_collapse}.
% , where the claim is that the directions of high variances are important. 
This requires further analysis and is beyond the scope of this work.

\section{Results}
\label{sec:results}

\subsection{Experiments on CityScapes}
\noindent\textbf{The size of the training set matters for linear models, less for k-NN. }
In \cref{fig:training-set-size} we illustrate the performance of k-NN and linear classifiers for varying sizes of training dataset. 
We first observe that the linear models generally achieve better results than k-NN, especially with more training data. 
However, for the DINO versions and the supervised ViT, k-NN and linear classifiers demonstrate comparable performance. 
Instead, MAE representations yield bad k-NN performance. While its linear results are only slightly inferior to its DINO counterpart, the k-NN classifier leads to a remarkable about 4 times worse performance.

\begin{figure*}[t]
    \centering
    \includegraphics[width=0.85\linewidth]{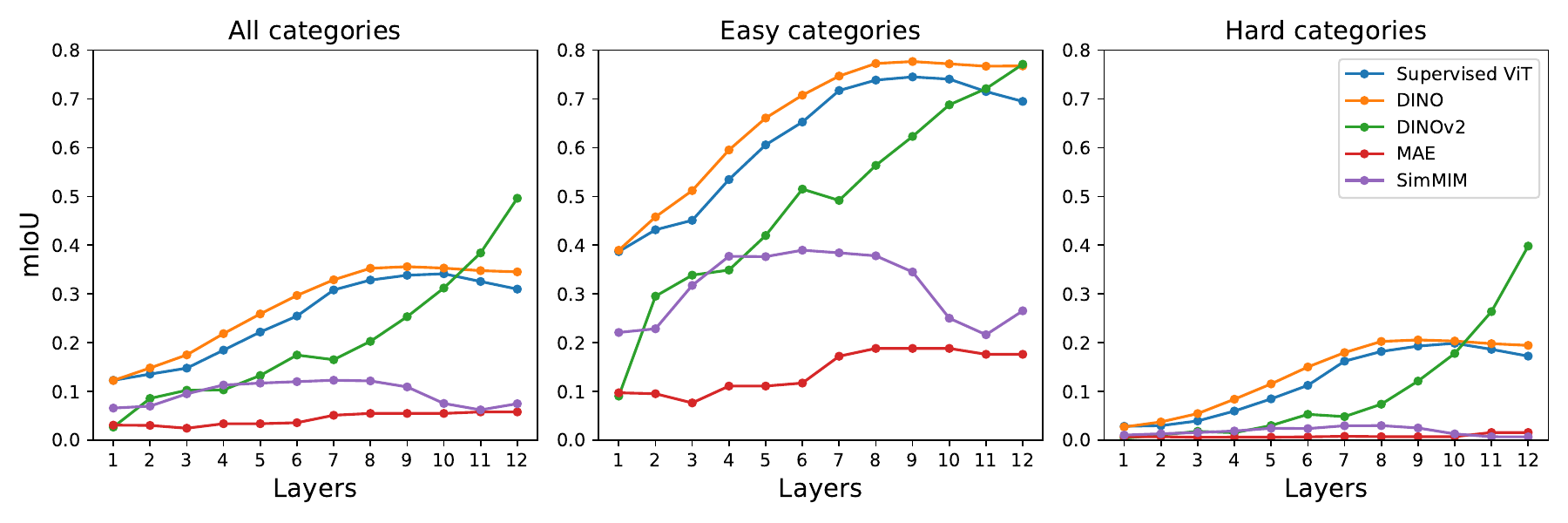}
    \caption{k-NN patch classification performance (mIoU) for various vision transformers on few-shot subset of Cityscapes.}
    \label{fig:layerwise-knn-cityscapes}
\end{figure*}

\noindent\textbf{Evaluating robustness to degradations.}
According to~\cite{DINOv2}, DINOv2 is extremely robust compared to other pretrained vision transformers as measured by its performance on domain-shifted versions of ImageNet. 
In this subsection, we perform such an analysis, but on the image patch representation level instead.

\begin{figure}[t]
    \centering
    \includegraphics[width=0.85\linewidth]{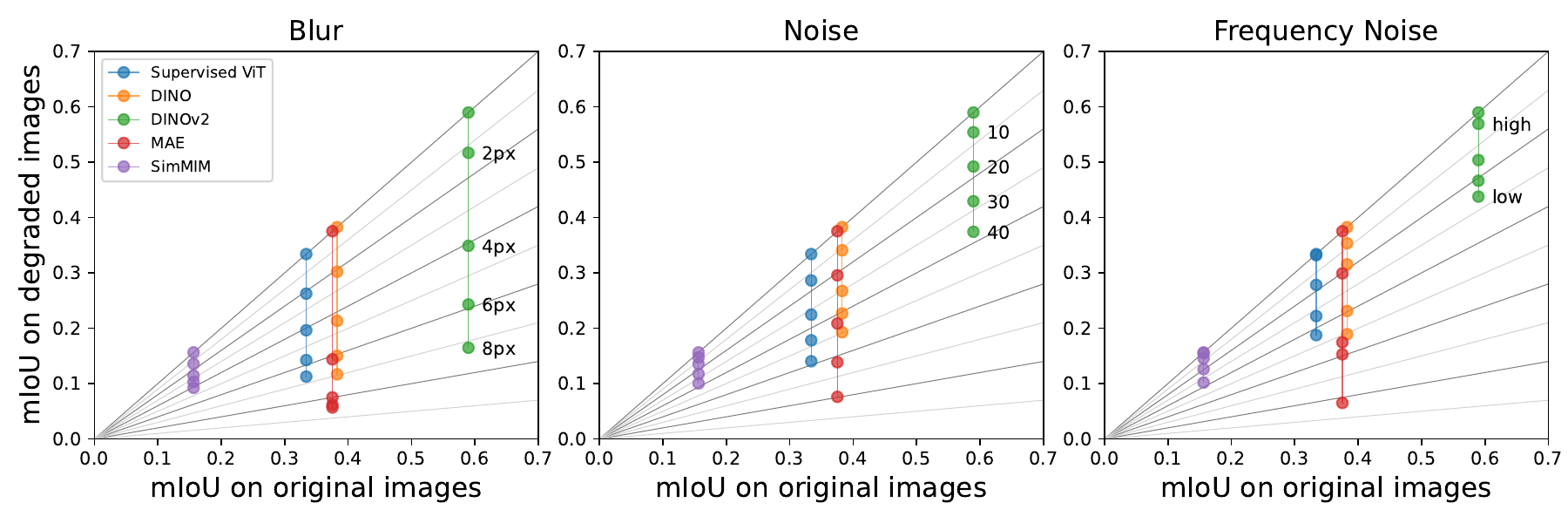}
    \caption{Robustness of linear models for segmentation on Cityscapes. For every model (differing by colors) we have five dots, each representing one level of degradation. y-axis shows the mIoU on corrupted images. The top dots in vertical lines correspond to the original images. Grey diagonal lines indicate levels of equal relative drop in mIoU.}
    \label{fig:segmentation-degradation}
\end{figure}

We applied three sets of corruptions to the same small subset of the validation set of Cityscapes. First, we blur out the images by applying four sizes of the blur kernel: $10 \times 10$, $20 \times 20$, $30 \times 30$ and $40 \times 40$. 
Second, we add four different levels of Gaussian noise (with mean $0$ and standard deviations $10, 20, 30,$ and $40$).
At last, we add frequency-based random noise (with $40$ standard deviation) to the images the following way. 
We generate a Gaussian noise of the same shape as the source image, convert it into the frequency domain, keep the frequencies only inside one of the four narrow bands, convert the noisy image back to image pixels, and add it to the original image. \cref{fig:frequency_based_random_noise} in Appendix visualizes this process.
We feed these degraded images to the ViTs and check the robustness of the methods for the different types and levels of degradations. The first row of \cref{fig:segmentation-degradation} demonstrates the results of k-NN for blurred, additive Gaussian noise and added frequency-based random noise for each color, respectively.

One can claim that the robustness to various degradations may be attributed to the augmentations used during training. 
Models that employ color enhancement augmentations are assumed to exhibit higher resilience against those types of degradations (blur, Gaussian noise, etc).
However, according to~\cite{MAE} color jittering-based augmentation degrades its performance.
This suggests a potential trade-off between performance and robustness to degradations.
Due to computational limitations, further investigation of this is left for future work.

\noindent\textbf{DINOv2 is more robust when tested on small degradations.} Fig.~\ref{fig:segmentation-degradation} shows that DINOv2 is relatively more robust for the smallest blur radius compared to DINO and Supervised ViT. For stronger blurred versions these three models degrade by almost the same ratio. MAE degrades relatively faster than others. k-NN results are similar to linear probing results, except for MAE, for which even the smallest degradation leads to a collapse in predictions: k-NN predicts the same class for all patches (usually \textit{vegetation} or \textit{sky}). 
\begin{figure}[!ht]
    \centering
    \begin{subfigure}[b]{0.98\textwidth}
        \centering
        \includegraphics[width=\textwidth]{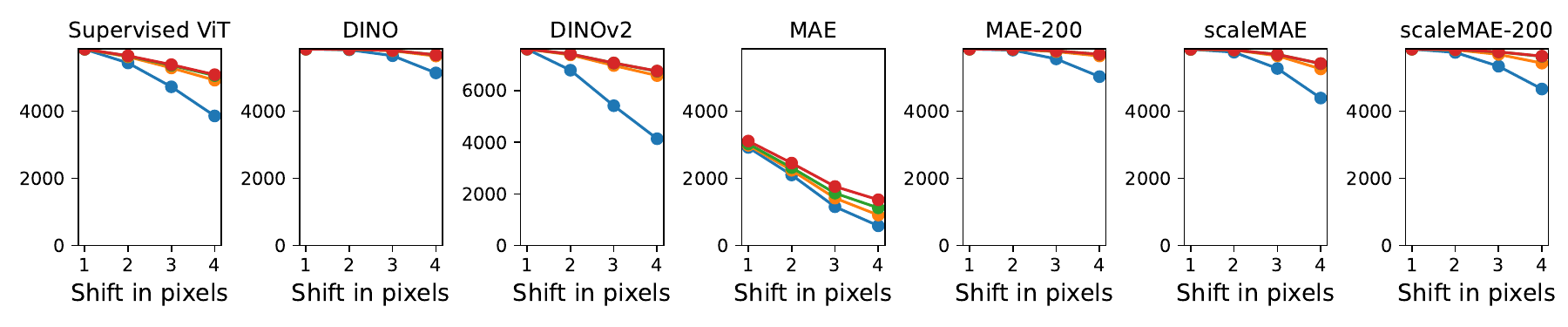}
        \caption{Shifts}
        \label{fig:fair1m-degradation-shifts}
    \end{subfigure}
    \hfill
     \begin{subfigure}[b]{0.98\textwidth}
        \centering
        \includegraphics[width=\textwidth]{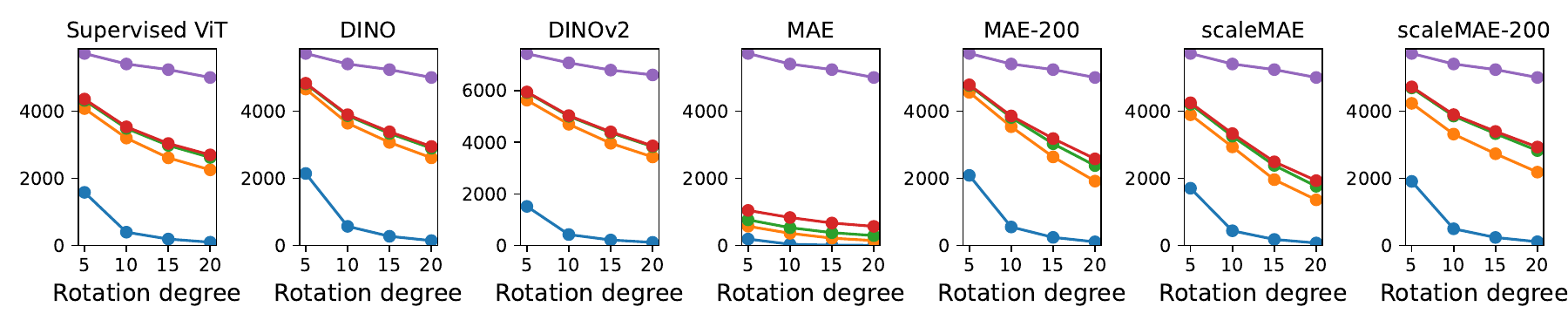}
        \caption{Rotation}
        \label{fig:fair1m-degradation-rotates}
    \end{subfigure}
    \begin{subfigure}[b]{0.98\textwidth}
        \centering
        \includegraphics[width=\textwidth]{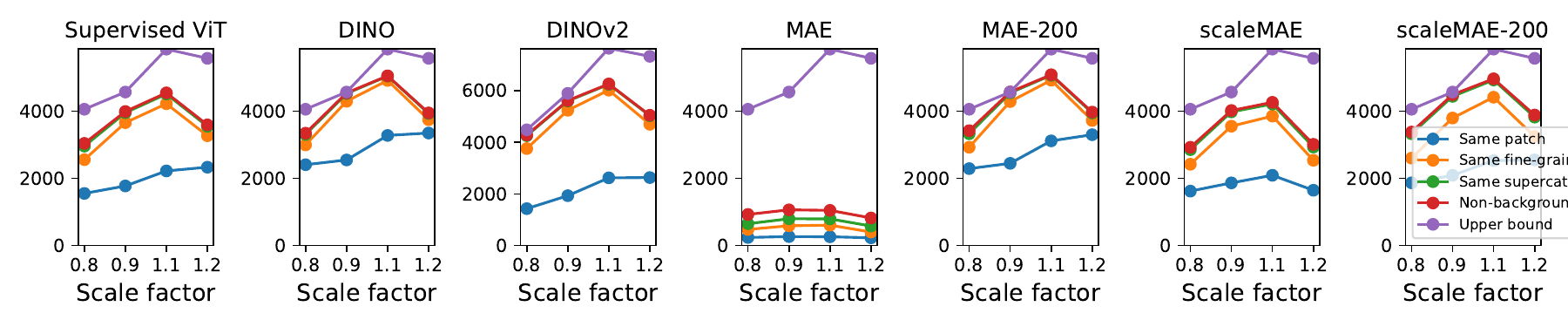}
        \caption{Scale}
        \label{fig:fair1m-degradation-scales}
    \end{subfigure}
    \hfill
    \begin{subfigure}[b]{0.98\textwidth}
    \centering
    \includegraphics[width=\textwidth]{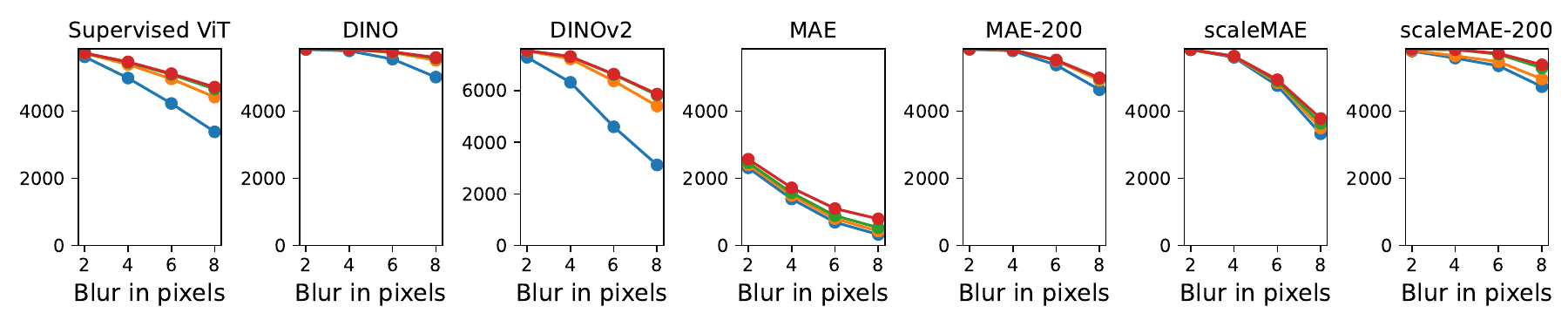}
    \caption{Blur}
    \label{fig:fair1m-degradation-blurs}
    \end{subfigure}
    \hfill
    \begin{subfigure}[b]{0.98\textwidth}
        \centering
        \includegraphics[width=\textwidth]{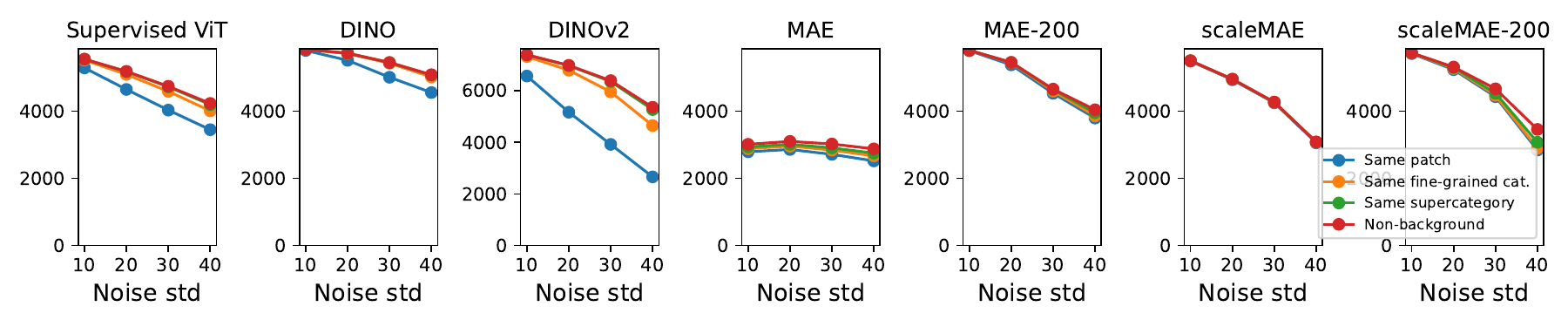}
        \caption{Gaussian noise}
        \label{fig:fair1m-degradation-noises}
    \end{subfigure}
    \hfill
    \caption{Stacked area plots for each model (columns) and transformation type (rows). In each subplot, x-axis is the level of transformation, y-axis is the percentage of correctly matched patches in each granularity.}
    \label{fig:fair1m-degradation-stacked-area}
\end{figure}
\noindent\textbf{Supervised ViT does not suffer from high-frequency noise.} According to~\cite{what-SS-ViT-learns-iclr2023},
masking-based methods like MAE rely more on high frequency features, while the methods based on contrastive training (including DINO) rely more on lower frequency features. This implies that DINO representations should be more robust wrt high frequency noise, while MAE representations should perform better under low frequency noise. In our experiments, MAE performs worse for all frequencies of noise. Instead, Supervised ViT is 100\% robust with respect to high frequency noise. 
This can be explained by its objective to learn object categories of the full image which makes its last layers to forget irrelevant high-frequency information. A similar phenomenon was reported in \cite{usable-information}.
% {\color{red} we need degradation analysis for MAE-200}

\noindent\textbf{The findings are confirmed on ADE20K.} We have created a similar few-shot subset of ADE20K training data which consists of 600 images in the training set (4 images per class) of size  672x448 and another 300 images in the validation set. With both k-NN and linear probing we get similar relative performance among the ViTs we have tested (\cref{tab:cs_ade20k_knn_linear} in \cref{sec:app-more-datasets}).

\subsection{Experiments on Fair1M.}
For the next set of experiments, we utilized the Fair1M dataset and generated transformed versions of all images. 
In the first set of experiments, we applied diagonal shifts by $1, 2, 3,$ and $4$ pixels. In the second set of experiments, we rotated the images by $5,$ $10,$ $15,$ and $20$ degrees counterclockwise. In the third set of experiments, we scaled the images by factors of $0.8$, $0.9$, $1.1$, and $1.2$. For the fourth and fifth sets of experiments, we applied blur and Gaussian noise degradations. We computed patch representations for all these images, creating many images with the same instances of objects.

For each patch of a transformed image covering an annotated object, we retrieve the closest patch from the complete set of original image patches. Ideally, the closest patch should be the original version of the patch (without blur, noise, or shift). 
Otherwise, the second best option should be another instance of the same fine-grained category, the third best option would be a patch of an object within the same supercategory. 
The worst-case scenario occurs when the closest patch belongs to another category or is a background patch.
For every model and image transformation level we calculate the number of patches for which the closest patch belongs to the mentioned classes.

\noindent\textbf{Increased levels of image degradation decrease all metrics.} The results are presented in \cref{fig:fair1m-degradation-stacked-area}. When the target patch transformation is small (e.g.\ Gaussian noise with just 10 pixels standard deviation), the closest patch is almost always the original one for all models. One notable exception is MAE, for which in case of roughly 40-50\% patches the closest one is the right one, but for the remaining ones the closest patch is a background patch. With stronger transformations, the ratio of correct patches decreases for all models, and more than half of the remaining patches are matched to patches of the same fine-grained object category (again, with the exception of MAE).

For the experiments involving image shifting, rotation, and scaling, the first level of evaluation, called 'same-patch' 
% is not straightforward because it is not immediately clear which patch corresponds to the original one. 
% We define the corresponding patch as the one containing the center point of the rotated patch. 
is challenging. 
We define the corresponding patch as the one containing the center point of the rotated patch.
We remark, that several patches near the corners of the rotated tile may lack a corresponding patch, placing an upper bound on same-patch retrieval accuracy. 
The results, along with the upper bounds, are shown in \cref{fig:fair1m-degradation-stacked-area}. 
We observe, that for rotation and scaling all models exhibit significantly less robustness even with a 5-degree rotation and a small amount of scaling compared to the highest blur radius or noise level attempted. 
The order of model performance remains consistent with other image transformations: DINO performs the best, followed by MAE (as well as Scale-MAE) with 200 high-variance features removed, followed by DINOv2 and Supervised ViT, with MAE as a distant outlier. 
It is worth noting that Scale-MAE outperforms MAE significantly; however, removing its 200 high-variance features also notably improves its performance.

\noindent\textbf{DINO is the most robust one.} Among the ViTs we tested, DINO is the most robust one across all transformations. DINOv2 is less robust by all metrics, and performs similar to Supervised ViT. 
To verify whether this disadvantage of DINOv2 comes from the patch-level loss term or the scale of the model and the dataset, we perform the same analysis with iBOT. iBOT performs slightly better than DINO (\cref{fig:fair1m-ibot} in Appendix), thus patch-level loss cannot be blamed.

\begin{figure}[t]
    \centering
    % \begin{subfigure}[b]{\textwidth}
        \centering
        \includegraphics[width=0.85\textwidth]{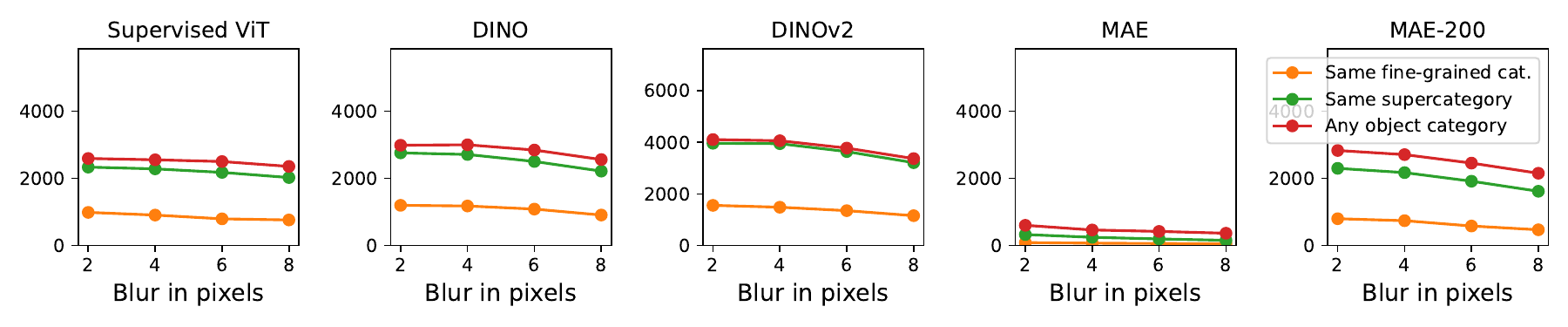}
    % \end{subfigure}
    % \hfill
    \caption{Patch retrieval results when the original image is not available. x-axis is the level of blur, y-axis is the number of correctly matched patches in each granularity.}
    \label{fig:fair1m-retrieval}
\end{figure}

\noindent\textbf{Retrieving patches from other images.} We notice that most of the retrieved patches come from the same image tile. 
One potential explanation is that patch embeddings carry image-level information.
% We repeat this experiment with the patches of the original image removed from the set of available patches. 
Repeating the experiment with original image patches excluded, the closest patch can now belong to the same fine-grained category, the same supercategory, an incorrect supercategory, or serve as a background patch.
Here DINOv2 takes the lead, Supervised ViT and DINO perform slightly worse, and MAE performs poorly (\cref{fig:fair1m-retrieval}).

\noindent\textbf{MAE's performance can be improved}. 
We repeated this experiment with MAE-200 and Scale-MAE-200. 
These versions of MAE and Scale-MAE perform significantly better than the original ones and outperform both DINOv2 and Supervised ViT, and fall short only of DINO. 
These results imply that removing the high-variance features of MAE embeddings not only helps in semantic segmentation, but also in the identification of specific object instances in transformed images. 
On the other hand, this adds to the evidence that the high-variance features do not hold any unique information necessary for identifying the instances, as for small transformations the performance is near ideal ~(\cref{fig:fair1m-degradation-stacked-area}).

\subsection{Tracking Experiments.}
\begin{figure}
    \centering
    % \vspace{-5ex}
    \includegraphics[width=0.4\textwidth]{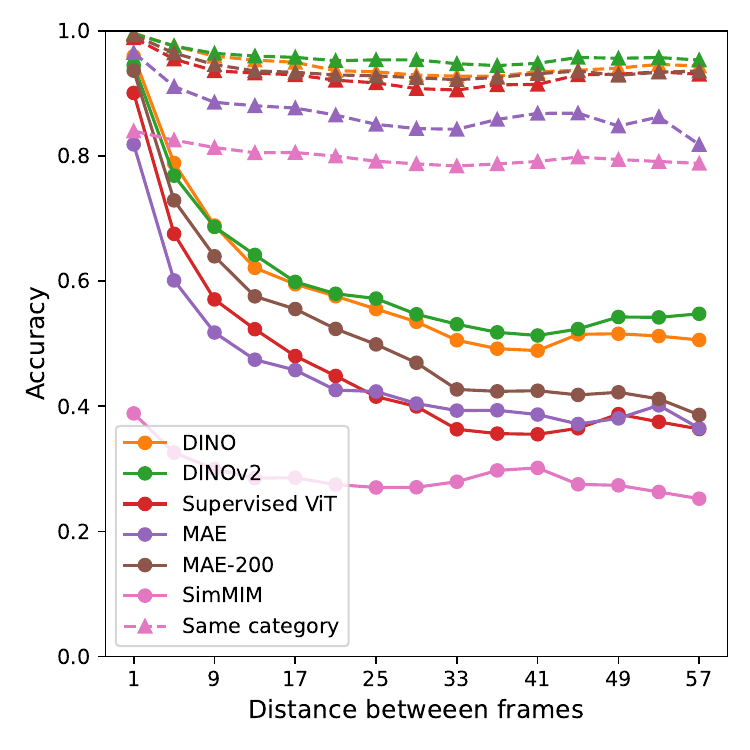}
    \caption{Object instance matching performance for various ViTs on the BDD100k MOT dataset. We report the ratio of correctly retrieved instances as a function of the frame gap $\Delta$. Solid lines with circles represent accuracy for instance retrieval, while dashed lines with triangles show the accuracy for retrieval within the same category.}
    % \vspace{-3ex}
    \label{fig:tracking}
\end{figure}

Next, we analyze the robustness of patch embeddings over time, as the objects undergo appearance changes.
We report the ratio of correctly retrieved instances and the percentage of object instances for which the retrieved instance has the same object category; see \cref{fig:tracking}.

% \begin{wrapfigure}{r}{0.45\textwidth}
% \end{wrapfigure}

\noindent\textbf{DINO-like models are best at tracking.} The accuracy of all models degrades for longer time intervals, as appearance changes in average increases over time. The retrieval performance stays relatively constant for intervals larger than one second, i.e.\ about 30 frames. DINO and DINOv2 show the best performance, with a slight advantage for DINO for $\Delta<10$, which is consistent with the patch retrieval experiments in \cref{sec:results}. Surprisingly, the Supervised ViT is substantially inferior to DINO, and performs slightly better than MAE.
We also notice that MAE-200 outperforms MAE across all values of $\Delta$, consistently surpasses Supervised ViT, and achieves comparable accuracy with DINO and DINOv2 for small values of $\Delta$. 
This indicates that MAE-200 improves the performance of tracking.

\noindent\textbf{Incorrect matches usually have the correct category.} When instance matching fails, the closest object belongs to the same object category in more than 95\% of cases for DINOv2, slightly less for DINO and Supervised ViT. MAE's performance is noticeably worse than Supervised ViT. These results are consistent with patch classification metrics from \cref{sec:results}.

We repeat the experiments on MOT17 dataset, observing a consistent pattern: contrastive approaches outperform ViTs based on MIM. Details in \cref{sec:app-more-datasets}.

\section{Discussion}
\label{sec:discussion}

\begin{figure}[t]
    \centering
    \includegraphics[width=0.85\linewidth]{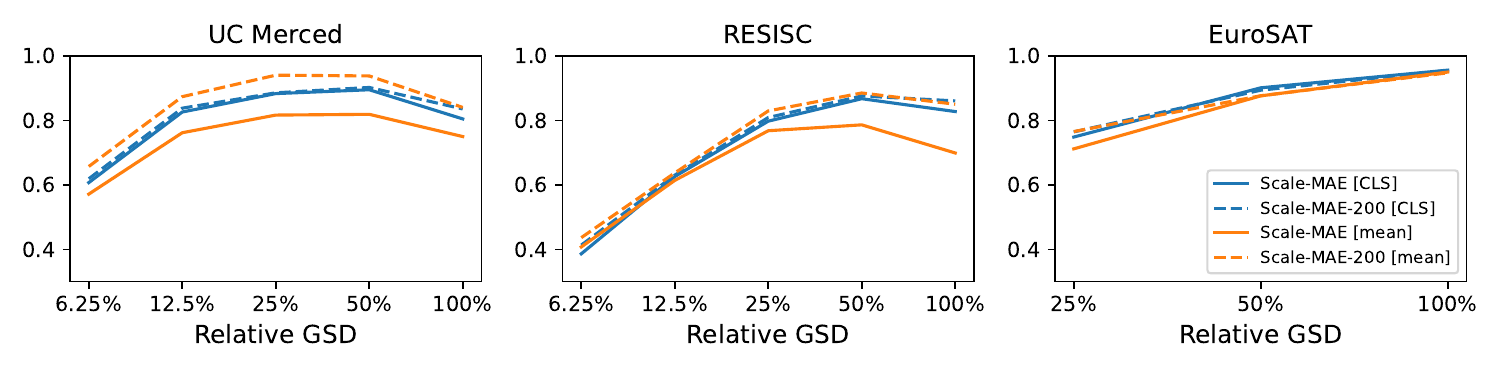}
    \caption{Comparison between the original Scale-MAE and our proposed variants, where we remove 200 high-variance features, using k-NN on three datasets for various GSDs.}
    \label{fig:scaleMAE_variance_experiments}
\end{figure}

\noindent\textbf{Remedy works for image-level tasks as well: ScaleMAE~\cite{scalemae}.} \\
% \noindent\textbf{Improving k-NN performance for Scale-MAE~\cite{scalemae}}. 
To validate our hypothesis, regarding high-variance features affecting distance-based metrics for MAE, we chose a
recent extension of MAE called Scale-MAE~\cite{scalemae}, specifically trained for satellite imagery.
Scale-MAE employs k-NN for performance comparison against state-of-the-art methods. We conducted experiments by removing its $m=200$ high-variance features.
% To test our hypothesis regarding high-variance features affecting distance-based metrics for MAE, we also use MAE-200 and Scale-MAE-200. These variants remove the 200 high-variance features from the representation vectors.
Scale-MAE is trained in a way that it has a meaningful [CLS] token, which can be used as a representation of an image along with the mean of patch tokens. The authors used [CLS] token representations for image classification using k-NN and showed that it outperforms MAE. We reproduced the results and confirmed that the mean of patch vectors is indeed worse than [CLS]. However, after removing the high-variance features, mean patch representations outperform [CLS] across almost all values of ground sampling distance (GSD) in the UCMerced and RESISC datasets. The results are presented in \cref{fig:scaleMAE_variance_experiments}. 
% Unlike the mean patch representations, we observe that removing 200 high-variance features from [CLS] does not change the results significantly.
Interestingly, the removal of 200 high-variance features from [CLS] does not significantly alter the outcomes.
% ; however, once we use the mean vectors across the image patches, and after removing 200 high-variance features from the patch representations, they outperform the proposed methods that utilize [CLS] representation.

% \subsection{What information do these features hold?}
\noindent\textbf{What information do these features hold?} \\
One hypothesis is that these features are necessary for identifying specific instances of objects among the same category, or for distinguishing fine-grained categories of objects. In \cref{sec:results} we provide negative evidence for this hypothesis: removing high-variance features improves retrieval performance so that the negative effect of losing some instance-specific information, if true, is not detectable.

\noindent\textbf{Image reconstruction quality is one metric that suffers.}
Another hypothesis is that this information is useful for pixel-level reconstruction of the patches. To verify this, we took MAE embeddings with $m=200$ features replaced with zeros, and used the pretrained decoder to reconstruct the image. %We evaluate reconstruction quality by Mean Square Error (MSE), Peak Signal to Noise Ratio (PSNR) and Structural Similarity Index (SSIM). 
Reconstruction error indeed increases with the replaced features (\cref{tab:mae-reconstruct} in Appendix). It is still unclear why the variance of those features is high.

\noindent\textbf{High variance features do not store global context.} Another hypothesis is that the high variance features might contain global information about the image not directly relevant to the semantic class of the current patch. An example of a global information is the number of pixels of each class in the entire image. To test whether patch representations contain such information, we use L2-regularized linear regression to predict those numbers (normalized by the total number of pixels) for the three most common classes (road, building, vegetation) from the representation of a central patch. We report $R^2$ for each class and average them. We see that MAE embeddings without top 200 high-variance features have the same predictive power (0.672) as the full MAE embeddings (0.676). It implies that the high variance features do not contain additional information about the global context not present in the other features. Furthermore, the features with the highest variances are less predictive (0.550) than a random subset of 200 features (0.666); see \cref{tab:r_squared} of Appendix.

\noindent\textbf{ViTs with no reconstruction loss do not have this phenomenon.}
We performed similar analyses for all other ViTs, and visualize the results on \cref{fig:nonhighvariance}. The phenomenon of improved k-NN performance when high-variance features are removed is present only in the models trained with pixel-level reconstruction objective. In case of other ViTs, while removing high-variance features does not improve k-NN performance, it does not harm it either. Linear probing performance also stays robust with these removals up to some level.

\noindent\textbf{Feature normalization has a similar effect for MAE, but not for SimMIM.} We created another version of MAE embeddings by applying the pretrained BatchNorm layer from the linear model, before passing them to k-NN classifier. It helped to improve k-NN performance by almost the same amount as removing $m=200$ features. It improved SimMIM's k-NN performance as well, but linear probing worsened significantly. We conclude that feature normalization is an alternative but not an identical strategy to feature removal.
% for minimizing the negative impact of high-variance features. 

\noindent\textbf{High variance features of MAE are stable across datasets.} We identify top 200 high-variance features of patch representations extracted from Cityscapes, ADE20K and FAIR1M datasets. 196 of these features are shared among Cityscapes and ADE20K, while 192 of them are shared among Cityscapes and FAIR1M.

To summarize, the high-variance features do not hold semantic or global information, they store some pixel-level details useful in image reconstruction, and are consistently detected across all MIM-based methods and datasets.

% \subsection{Distinct layer-wise behavior across ViTs. }
\noindent\textbf{Distinct layer-wise behavior across ViTs. } \\
We perform a detailed analysis of the k-NN patch classification performance for representations extracted at different layers of the network in \cref{fig:layerwise-knn-cityscapes}. Supervised ViT and DINO perform strikingly similarly. The performance slowly improves from the first layer to the 8th and then saturates. There is a slight drop in performance in the last two layers, which is more noticeable in the case of Supervised ViT. We remark that a similar approach has been observed in~\cite{gfm}.

DINOv2's behavior is quite different. In the first layers its performance is worse than DINO. For the easiest five object categories (\textit{road}, \textit{vegetation}, \textit{sky}, \textit{car}, \textit{building}) the performance catches up in the last layer. 
For harder object categories DINOv2 catches up with DINO and Supervised ViT in the 10th layer, then significantly outperforms them in the 11th and 12th layers. 
In particular, IoU of \textit{bus} category jumps from 0.059 on the 9th layer to 0.729 on the 12th layer. 
The advantage of DINOv2 thus mostly stem from the harder categories. 
Again, we observe poor k-NN performance in case of the MAE. 
SimMIM performs better than MAE, but only in the middle layers. The difference is more significant for easier object categories. The quality of the last three layers is similar to MAE.

\section{Conclusion and Limitations}
We perform a comprehensive analysis and comparison of the quality and properties of locality patch embeddings extracted from self-supervised ViT models. We observe that the contrastive learning based DINO series outperform both supervised and masked image modeling approaches. Moreover, we identify and study the inferior k-NN classification performance of MAE which limits its use without fine-tuning. We find that the features with relatively high variances are not informative for patch classification or retrieval tasks and that their removal results in improvement for k-NN performance while not harming linear probing. 

\noindent\noindent\textbf{Limitations. } Due to the high computational cost, we are unable to retrain networks, preventing us from analyzing architectural choices or loss components in the discussed Vision Transformers (ViTs). Therefore, our comparison is restricted to existing pre-trained networks.

\subsubsection*{Acknowledgements}
We would like to thank Hrayr Harutyunyan and Armen Aghajanyan for helpful comments. Part of this work was supported by CSIE Foundation. Vahan Huroyan's work is supported by RA Higher Education and Science Committee, in the frames of the research project No. 23PostDoc-1B009.

% \newpage


{\small
\bibliographystyle{ieeenat_fullname}
\bibliography{vision}

\begin{thebibliography}{35}
\providecommand{\natexlab}[1]{#1}
\providecommand{\url}[1]{\texttt{#1}}
\expandafter\ifx\csname urlstyle\endcsname\relax
  \providecommand{\doi}[1]{doi: #1}\else
  \providecommand{\doi}{doi: \begingroup \urlstyle{rm}\Url}\fi

\bibitem[Ba et~al.(2016)Ba, Kiros, and Hinton]{layer-norm}
Jimmy~Lei Ba, Jamie~Ryan Kiros, and Geoffrey~E Hinton.
\newblock Layer normalization.
\newblock \emph{arXiv preprint arXiv:1607.06450}, 2016.

\bibitem[Bhojanapalli et~al.(2021)Bhojanapalli, Chakrabarti, Glasner, Li, Unterthiner, and Veit]{bhojanapalli2021understanding}
Srinadh Bhojanapalli, Ayan Chakrabarti, Daniel Glasner, Daliang Li, Thomas Unterthiner, and Andreas Veit.
\newblock Understanding robustness of transformers for image classification.
\newblock In \emph{2021 {IEEE/CVF} International Conference on Computer Vision, {ICCV} 2021, Montreal, QC, Canada, October 10-17, 2021}, pages 10211--10221. {IEEE}, 2021.

\bibitem[Brown et~al.(2020)Brown, Mann, Ryder, Subbiah, Kaplan, Dhariwal, Neelakantan, Shyam, Sastry, Askell, et~al.]{GPT3}
Tom Brown, Benjamin Mann, Nick Ryder, Melanie Subbiah, Jared~D Kaplan, Prafulla Dhariwal, Arvind Neelakantan, Pranav Shyam, Girish Sastry, Amanda Askell, et~al.
\newblock Language models are few-shot learners.
\newblock \emph{Advances in neural information processing systems}, 33:\penalty0 1877--1901, 2020.

\bibitem[Caron et~al.(2021)Caron, Touvron, Misra, J{\'{e}}gou, Mairal, Bojanowski, and Joulin]{DINO}
Mathilde Caron, Hugo Touvron, Ishan Misra, Herv{\'{e}} J{\'{e}}gou, Julien Mairal, Piotr Bojanowski, and Armand Joulin.
\newblock Emerging properties in self-supervised vision transformers.
\newblock In \emph{2021 {IEEE/CVF} International Conference on Computer Vision, {ICCV} 2021, Montreal, QC, Canada, October 10-17, 2021}, pages 9630--9640. {IEEE}, 2021.

\bibitem[Cheng et~al.(2022)Cheng, Misra, Schwing, Kirillov, and Girdhar]{Mask2Former}
Bowen Cheng, Ishan Misra, Alexander~G. Schwing, Alexander Kirillov, and Rohit Girdhar.
\newblock Masked-attention mask transformer for universal image segmentation.
\newblock In \emph{{IEEE/CVF} Conference on Computer Vision and Pattern Recognition, {CVPR} 2022, New Orleans, LA, USA, June 18-24, 2022}, pages 1280--1289. {IEEE}, 2022.

\bibitem[Cordts et~al.(2016)Cordts, Omran, Ramos, Rehfeld, Enzweiler, Benenson, Franke, Roth, and Schiele]{cityscapes}
Marius Cordts, Mohamed Omran, Sebastian Ramos, Timo Rehfeld, Markus Enzweiler, Rodrigo Benenson, Uwe Franke, Stefan Roth, and Bernt Schiele.
\newblock The cityscapes dataset for semantic urban scene understanding.
\newblock In \emph{Proceedings of the IEEE Conference on Computer Vision and Pattern Recognition (CVPR)}, 2016.

\bibitem[Cubuk et~al.(2020)Cubuk, Zoph, Shlens, and Le]{randaugment}
Ekin~Dogus Cubuk, Barret Zoph, Jonathon Shlens, and Quoc Le.
\newblock Randaugment: Practical automated data augmentation with a reduced search space.
\newblock In \emph{Advances in Neural Information Processing Systems 33: Annual Conference on Neural Information Processing Systems 2020, NeurIPS 2020, December 6-12, 2020, virtual}, 2020.

\bibitem[Darcet et~al.(2023)Darcet, Oquab, Mairal, and Bojanowski]{registers}
Timoth{\'e}e Darcet, Maxime Oquab, Julien Mairal, and Piotr Bojanowski.
\newblock Vision transformers need registers.
\newblock \emph{arXiv preprint arXiv:2309.16588}, 2023.

\bibitem[Dehghani et~al.(2023)Dehghani, Djolonga, Mustafa, Padlewski, Heek, Gilmer, Steiner, Caron, Geirhos, Alabdulmohsin, Jenatton, Beyer, Tschannen, Arnab, Wang, Ruiz, Minderer, Puigcerver, Evci, Kumar, van Steenkiste, Elsayed, Mahendran, Yu, Oliver, Huot, Bastings, Collier, Gritsenko, Birodkar, Vasconcelos, Tay, Mensink, Kolesnikov, Pavetic, Tran, Kipf, Lucic, Zhai, Keysers, Harmsen, and Houlsby]{vit22b}
Mostafa Dehghani, Josip Djolonga, Basil Mustafa, Piotr Padlewski, Jonathan Heek, Justin Gilmer, Andreas~Peter Steiner, Mathilde Caron, Robert Geirhos, Ibrahim Alabdulmohsin, Rodolphe Jenatton, Lucas Beyer, Michael Tschannen, Anurag Arnab, Xiao Wang, Carlos~Riquelme Ruiz, Matthias Minderer, Joan Puigcerver, Utku Evci, Manoj Kumar, Sjoerd van Steenkiste, Gamaleldin~Fathy Elsayed, Aravindh Mahendran, Fisher Yu, Avital Oliver, Fantine Huot, Jasmijn Bastings, Mark Collier, Alexey~A. Gritsenko, Vighnesh Birodkar, Cristina~Nader Vasconcelos, Yi Tay, Thomas Mensink, Alexander Kolesnikov, Filip Pavetic, Dustin Tran, Thomas Kipf, Mario Lucic, Xiaohua Zhai, Daniel Keysers, Jeremiah~J. Harmsen, and Neil Houlsby.
\newblock Scaling vision transformers to 22 billion parameters.
\newblock In \emph{International Conference on Machine Learning, {ICML} 2023}, pages 7480--7512. {PMLR}, 2023.

\bibitem[Doersch et~al.(2015)Doersch, Gupta, and Efros]{context_predictor}
Carl Doersch, Abhinav Gupta, and Alexei~A. Efros.
\newblock Unsupervised visual representation learning by context prediction.
\newblock In \emph{2015 {IEEE} International Conference on Computer Vision, {ICCV} 2015, Santiago, Chile, December 7-13, 2015}, pages 1422--1430. {IEEE} Computer Society, 2015.

\bibitem[Dosovitskiy et~al.(2021)Dosovitskiy, Beyer, Kolesnikov, Weissenborn, Zhai, Unterthiner, Dehghani, Minderer, Heigold, Gelly, Uszkoreit, and Houlsby]{ViT}
Alexey Dosovitskiy, Lucas Beyer, Alexander Kolesnikov, Dirk Weissenborn, Xiaohua Zhai, Thomas Unterthiner, Mostafa Dehghani, Matthias Minderer, Georg Heigold, Sylvain Gelly, Jakob Uszkoreit, and Neil Houlsby.
\newblock An image is worth 16x16 words: Transformers for image recognition at scale.
\newblock In \emph{9th International Conference on Learning Representations, {ICLR} 2021, Virtual Event, Austria, May 3-7, 2021}. OpenReview.net, 2021.

\bibitem[Garcia et~al.(2023)Garcia, Bansal, Cherry, Foster, Krikun, Johnson, and Firat]{unreasonable-machine-translation}
Xavier Garcia, Yamini Bansal, Colin Cherry, George Foster, Maxim Krikun, Melvin Johnson, and Orhan Firat.
\newblock The unreasonable effectiveness of few-shot learning for machine translation.
\newblock In \emph{International Conference on Machine Learning}, pages 10867--10878. PMLR, 2023.

\bibitem[He et~al.(2017)He, Gkioxari, Doll{\'{a}}r, and Girshick]{mask-rcnn}
Kaiming He, Georgia Gkioxari, Piotr Doll{\'{a}}r, and Ross~B. Girshick.
\newblock Mask {R-CNN}.
\newblock In \emph{{IEEE} International Conference on Computer Vision, {ICCV} 2017, Venice, Italy, October 22-29, 2017}, pages 2980--2988. {IEEE} Computer Society, 2017.

\bibitem[He et~al.(2020)He, Fan, Wu, Xie, and Girshick]{MOCO}
Kaiming He, Haoqi Fan, Yuxin Wu, Saining Xie, and Ross~B. Girshick.
\newblock Momentum contrast for unsupervised visual representation learning.
\newblock In \emph{2020 {IEEE/CVF} Conference on Computer Vision and Pattern Recognition, {CVPR} 2020, Seattle, WA, USA, June 13-19, 2020}, pages 9726--9735. Computer Vision Foundation / {IEEE}, 2020.

\bibitem[He et~al.(2022)He, Chen, Xie, Li, Doll{\'{a}}r, and Girshick]{MAE}
Kaiming He, Xinlei Chen, Saining Xie, Yanghao Li, Piotr Doll{\'{a}}r, and Ross~B. Girshick.
\newblock Masked autoencoders are scalable vision learners.
\newblock In \emph{{IEEE/CVF} Conference on Computer Vision and Pattern Recognition, {CVPR} 2022, New Orleans, LA, USA, June 18-24, 2022}, pages 15979--15988. {IEEE}, 2022.

\bibitem[Jing et~al.(2022)Jing, Vincent, LeCun, and Tian]{dimensional_collapse}
Li Jing, Pascal Vincent, Yann LeCun, and Yuandong Tian.
\newblock Understanding dimensional collapse in contrastive self-supervised learning.
\newblock In \emph{The Tenth International Conference on Learning Representations, {ICLR} 2022, Virtual Event, April 25-29, 2022}. OpenReview.net, 2022.

\bibitem[Kleinman et~al.(2021)Kleinman, Achille, Idnani, and Kao]{usable-information}
Michael Kleinman, Alessandro Achille, Daksh Idnani, and Jonathan~C. Kao.
\newblock Usable information and evolution of optimal representations during training.
\newblock In \emph{9th International Conference on Learning Representations, {ICLR} 2021, Virtual Event, Austria, May 3-7, 2021}. OpenReview.net, 2021.

\bibitem[Mendieta et~al.(2023)Mendieta, Han, Shi, Zhu, and Chen]{gfm}
Mat{\'{\i}}as Mendieta, Boran Han, Xingjian Shi, Yi Zhu, and Chen Chen.
\newblock Towards geospatial foundation models via continual pretraining.
\newblock In \emph{{IEEE/CVF} International Conference on Computer Vision, {ICCV} 2023, Paris, France, October 1-6, 2023}, pages 16760--16770. {IEEE}, 2023.

\bibitem[Naseer et~al.(2021)Naseer, Ranasinghe, Khan, Hayat, Khan, and Yang]{naseer2021intriguing}
Muzammal Naseer, Kanchana Ranasinghe, Salman Khan, Munawar Hayat, Fahad~Shahbaz Khan, and Ming{-}Hsuan Yang.
\newblock Intriguing properties of vision transformers.
\newblock In \emph{Advances in Neural Information Processing Systems 34: Annual Conference on Neural Information Processing Systems 2021, NeurIPS 2021, December 6-14, 2021, virtual}, pages 23296--23308, 2021.

\bibitem[OpenAI(2023)]{GPT4}
OpenAI.
\newblock {GPT-4} technical report.
\newblock \emph{CoRR}, abs/2303.08774, 2023.

\bibitem[Oquab et~al.(2023)Oquab, Darcet, Moutakanni, Vo, Szafraniec, Khalidov, Fernandez, Haziza, Massa, El{-}Nouby, Assran, Ballas, Galuba, Howes, Huang, Li, Misra, Rabbat, Sharma, Synnaeve, Xu, J{\'{e}}gou, Mairal, Labatut, Joulin, and Bojanowski]{DINOv2}
Maxime Oquab, Timoth{\'{e}}e Darcet, Th{\'{e}}o Moutakanni, Huy Vo, Marc Szafraniec, Vasil Khalidov, Pierre Fernandez, Daniel Haziza, Francisco Massa, Alaaeldin El{-}Nouby, Mahmoud Assran, Nicolas Ballas, Wojciech Galuba, Russell Howes, Po{-}Yao Huang, Shang{-}Wen Li, Ishan Misra, Michael~G. Rabbat, Vasu Sharma, Gabriel Synnaeve, Hu Xu, Herv{\'{e}} J{\'{e}}gou, Julien Mairal, Patrick Labatut, Armand Joulin, and Piotr Bojanowski.
\newblock Dinov2: Learning robust visual features without supervision.
\newblock \emph{CoRR}, abs/2304.07193, 2023.

\bibitem[Park and Kim(2022)]{park2022how}
Namuk Park and Songkuk Kim.
\newblock How do vision transformers work?
\newblock In \emph{The Tenth International Conference on Learning Representations, {ICLR} 2022, Virtual Event, April 25-29, 2022}. OpenReview.net, 2022.

\bibitem[Park et~al.(2023)Park, Kim, Heo, Kim, and Yun]{what-SS-ViT-learns-iclr2023}
Namuk Park, Wonjae Kim, Byeongho Heo, Taekyung Kim, and Sangdoo Yun.
\newblock What do self-supervised vision transformers learn?
\newblock In \emph{The Eleventh International Conference on Learning Representations}, 2023.

\bibitem[Paul and Chen(2022)]{paul2022vision}
Sayak Paul and Pin{-}Yu Chen.
\newblock Vision transformers are robust learners.
\newblock In \emph{Thirty-Sixth {AAAI} Conference on Artificial Intelligence, {AAAI} 2022, Thirty-Fourth Conference on Innovative Applications of Artificial Intelligence, {IAAI} 2022, The Twelveth Symposium on Educational Advances in Artificial Intelligence, {EAAI} 2022 Virtual Event, February 22 - March 1, 2022}, pages 2071--2081. {AAAI} Press, 2022.

\bibitem[Radford et~al.(2021)Radford, Kim, Hallacy, Ramesh, Goh, Agarwal, Sastry, Askell, Mishkin, Clark, Krueger, and Sutskever]{CLIP}
Alec Radford, Jong~Wook Kim, Chris Hallacy, Aditya Ramesh, Gabriel Goh, Sandhini Agarwal, Girish Sastry, Amanda Askell, Pamela Mishkin, Jack Clark, Gretchen Krueger, and Ilya Sutskever.
\newblock Learning transferable visual models from natural language supervision.
\newblock In \emph{Proceedings of the 38th International Conference on Machine Learning, {ICML} 2021, 18-24 July 2021, Virtual Event}, pages 8748--8763. {PMLR}, 2021.

\bibitem[Raghu et~al.(2021)Raghu, Unterthiner, Kornblith, Zhang, and Dosovitskiy]{vits_see_like_cnn}
Maithra Raghu, Thomas Unterthiner, Simon Kornblith, Chiyuan Zhang, and Alexey Dosovitskiy.
\newblock Do vision transformers see like convolutional neural networks?
\newblock In \emph{Advances in Neural Information Processing Systems 34: Annual Conference on Neural Information Processing Systems 2021, NeurIPS 2021, December 6-14, 2021, virtual}, pages 12116--12128, 2021.

\bibitem[Reed et~al.(2022)Reed, Gupta, Li, Brockman, Funk, Clipp, Keutzer, Candido, Uyttendaele, and Darrell]{scalemae}
Colorado~J. Reed, Ritwik Gupta, Shufan Li, Sarah Brockman, Christopher Funk, Brian Clipp, Kurt Keutzer, Salvatore Candido, Matt Uyttendaele, and Trevor Darrell.
\newblock Scale-mae: {A} scale-aware masked autoencoder for multiscale geospatial representation learning.
\newblock \emph{CoRR}, abs/2212.14532, 2022.

\bibitem[Steiner et~al.(2021)Steiner, Kolesnikov, Zhai, Wightman, Uszkoreit, and Beyer]{vit_train}
Andreas Steiner, Alexander Kolesnikov, Xiaohua Zhai, Ross Wightman, Jakob Uszkoreit, and Lucas Beyer.
\newblock How to train your vit? data, augmentation, and regularization in vision transformers.
\newblock \emph{CoRR}, abs/2106.10270, 2021.

\bibitem[Sun et~al.(2022)Sun, Wang, Yan, Xu, Wang, Diao, Chen, Li, Feng, Xu, Weinmann, Hinz, Wang, and Fu]{sun2021fair1m}
Xian Sun, Peijin Wang, Zhiyuan Yan, Feng Xu, Ruiping Wang, Wenhui Diao, Jin Chen, Jihao Li, Yingchao Feng, Tao Xu, Martin Weinmann, Stefan Hinz, Cheng Wang, and Kun Fu.
\newblock Fair1m: A benchmark dataset for fine-grained object recognition in high-resolution remote sensing imagery.
\newblock \emph{ISPRS Journal of Photogrammetry and Remote Sensing}, 184:\penalty0 116--130, 2022.

\bibitem[Wang et~al.(2004)Wang, Bovik, Sheikh, and Simoncelli]{ssim}
Zhou Wang, Alan~C. Bovik, Hamid~R. Sheikh, and Eero~P. Simoncelli.
\newblock Image quality assessment: from error visibility to structural similarity.
\newblock \emph{{IEEE} Trans. Image Process.}, 13\penalty0 (4):\penalty0 600--612, 2004.

\bibitem[Xie et~al.(2022)Xie, Zhang, Cao, Lin, Bao, Yao, Dai, and Hu]{SimMIM}
Zhenda Xie, Zheng Zhang, Yue Cao, Yutong Lin, Jianmin Bao, Zhuliang Yao, Qi Dai, and Han Hu.
\newblock Simmim: a simple framework for masked image modeling.
\newblock In \emph{{IEEE/CVF} Conference on Computer Vision and Pattern Recognition, {CVPR} 2022, New Orleans, LA, USA, June 18-24, 2022}, pages 9643--9653. {IEEE}, 2022.

\bibitem[Yu et~al.(2020)Yu, Chen, Wang, Xian, Chen, Liu, Madhavan, and Darrell]{bdd100k}
Fisher Yu, Haofeng Chen, Xin Wang, Wenqi Xian, Yingying Chen, Fangchen Liu, Vashisht Madhavan, and Trevor Darrell.
\newblock {BDD100K:} {A} diverse driving dataset for heterogeneous multitask learning.
\newblock In \emph{2020 {IEEE/CVF} Conference on Computer Vision and Pattern Recognition, {CVPR} 2020, Seattle, WA, USA, June 13-19, 2020}, pages 2633--2642. Computer Vision Foundation / {IEEE}, 2020.

\bibitem[Yu et~al.(2023)Yu, Shi, Pasunuru, Muller, Golovneva, Wang, Babu, Tang, Karrer, Sheynin, et~al.]{CM3leon}
Lili Yu, Bowen Shi, Ramakanth Pasunuru, Benjamin Muller, Olga Golovneva, Tianlu Wang, Arun Babu, Binh Tang, Brian Karrer, Shelly Sheynin, et~al.
\newblock Scaling autoregressive multi-modal models: Pretraining and instruction tuning.
\newblock \emph{arXiv preprint arXiv:2309.02591}, 2023.

\bibitem[Zhang et~al.(2018)Zhang, Ciss{\'{e}}, Dauphin, and Lopez{-}Paz]{mixup}
Hongyi Zhang, Moustapha Ciss{\'{e}}, Yann~N. Dauphin, and David Lopez{-}Paz.
\newblock mixup: Beyond empirical risk minimization.
\newblock In \emph{6th International Conference on Learning Representations, {ICLR} 2018, Vancouver, BC, Canada, April 30 - May 3, 2018, Conference Track Proceedings}. OpenReview.net, 2018.

\bibitem[Zhou et~al.(2021)Zhou, Wei, Wang, Shen, Xie, Yuille, and Kong]{ibot}
Jinghao Zhou, Chen Wei, Huiyu Wang, Wei Shen, Cihang Xie, Alan~L. Yuille, and Tao Kong.
\newblock ibot: Image {BERT} pre-training with online tokenizer.
\newblock \emph{CoRR}, abs/2111.07832, 2021.

\end{thebibliography}
}

\ifarxiv \clearpage \appendix \section{Appendix}
\label{sec:appendix_section}

\subsection{The Choice of Self-supervised Vision Transformers}\label{sec:app-choice-of-vits}

We use the following pretrained visual transformers in our analysis.

\textbf{DINO}~\cite{DINO} is a self-supervised ViT that utilizes a self-distillation (student-teacher) framework. Different augmented versions of the same image pass through the teacher and student networks, and the student network is optimized to produce the same [CLS] vector as the teacher. Teacher's weights are then updated from the student's weight using exponential moving average.

\textbf{Masked Autoencoder (MAE)}~\cite{MAE} is trained to reconstruct the original image given partial observations. During training, a large random fraction of image patches are masked out in the input. The encoder is applied to the visible patches only. A relatively lightweight decoder gets encoder's outputs as input, along with [MASK] tokens for the masked patches, and attempts to reconstruct the original image. We use the pretrained encoder as a feature extractor for patches.

\textbf{SimMIM}~\cite{SimMIM} is another framework for vision transformers that uses masked image modeling. The main difference from MAE is that SimMIM uses a simple linear decoder on top of encoder's outputs.

In one experiment we analyzed \textbf{iBOT}~\cite{ibot}, which is another teacher-student framework that additionally masks some of the patches for the student network. In addition to DINO's objective, it has another loss term that forces the student network to produce patch representations for the masked patches similar to ones given by the teacher on an unmasked image. 

\textbf{DINOv2}~\cite{DINOv2} is a recent extension of iBOT which was trained on a much larger dataset. The dataset is comprised of 17 million images from ImageNet-2, Mapillary SLS and Google Landmarks v2, and additional 125 million images retrieved from a large pool of web-crawled images with the condition of being similar to images to a pre-selected 27 publicly available datasets. The main model has more than 1B parameters, which forced the authors to use multiple regularization techniques to stabilize the training. They also provide the distilled versions of the main model, which we use in our work.

Finally, we used a supervised baseline trained on ImageNet-1k with image-level labels. A linear layer was trained on top of the [CLS] token. Throughout the paper this one will be called   
\textbf{Supervised ViT}~\cite{ViT}. 
\footnote{
We used a version from \textit{torchvision} package: 
{\url{https://pytorch.org/vision/0.15/models/generated/torchvision.models.vit_b_16.html}}.
} 

All methods have been applied to multiple sizes of ViTs. In this work we focus only on one size that is available for all methods: ViT-B/16 with 86 million parameters. DINOv2 is the only one that does not have a ViT-B/16 version. Instead, we used the closest one: ViT-B/14, which is distilled from the ViT-g/14 model. This is another distinction between DINOv2 and others: the patches are a little bit smaller, and an image of size 224x224px has a larger number of DINOv2 patches.

The models also differ by the types of data augmentation used during pretraining. MAE used only simple resized crops and flips. DINO additionally used color jitter and blur with some differences between teacher and student networks. Supervised ViT uses a bunch of tricks as part of RandAugment~\cite{randaugment}, and also uses Mixup~\cite{mixup}. DINOv2's augmentations are similar to DINO. 
More details are available in \cref{tab:data_aug}.

We pass images through these ViTs and extract all patch embeddings from the 12-th layer. All ViTs apply layer normalization~\cite{layer-norm} on top of these embeddings. For consistency, we also apply layer normalization when we extract embeddings from the inner layers of ViTs.

% Throughout this paper, we will use representations or embeddings to refer to the feature vector extracted by the transformer at a given patch.

\subsection{Data Augmentations used in ViTs}

All ViTs we tested used data augmentation during the pre-training phase. In this section, we discuss the differences of augmentation strategies used.

In \textbf{DINO}~\cite{DINO} and \textbf{DINOv2}~\cite{DINOv2}  an image is cropped to two global crops or views for teacher network and multiple local views for student network. They apply different augmentations for different views. \textbf{MAE}~\cite{MAE} applies cropping-only augmentations. See the Table~\ref{tab:data_aug} for more details. For resized crop all models choose \(224\) for output size.

\textbf{Supervised ViT}~\cite{ViT} combines following techniques for data augmentation following to \cite{vit_train}
\begin{itemize}
    \item Mixup~\cite{mixup} with \(\alpha=0.2\), where \(\alpha=0\) means no Mixup.
    \item TensorFlow impementation of RandAugment~\cite{randaugment} with magnitude parameter \(m=15\) and number of augmentation layers \(l=2\).  
\end{itemize}
In~\cref{tab:data_aug}  we summarize the augmentation details.

\begin{table}[h]

\begin{tabular}{llll}
\toprule
\multicolumn{1}{l}{}                 &
\begin{tabular}[c]{c@{}@{}}
    \textbf{DINO}
\end{tabular}   
 &
\begin{tabular}[c]{c@{}@{}}
     \textbf{DINOv2}  
\end{tabular}
 &
\begin{tabular}[c]{c@{}@{}}
    \textbf{MAE} 
\end{tabular}                                                      \\ 
\midrule 
\multicolumn{1}{l}{\textbf{Resized Crop}}    & \begin{tabular}[l]{@{}l@{}}  \textbf{\textit{For global views}} \\
\hspace{0.5cm}\(scale=(0.4, 1.0)\)\\ \textbf{\textit{For local views}}\\
\hspace{0.5cm}\(scale=(0.05, 0.4)\)\end{tabular} & \begin{tabular}[l]{@{}l@{}}  \textbf{\textit{For global views}} \\ 
\hspace{0.5cm}\(scale=(0.32, 1.0)\)\\ \textbf{\textit{For local views}}\\
\hspace{0.5cm}\(scale=(0.05, 0.4)\)\end{tabular} & \begin{tabular}[l]{@{}l@{}}\\\\\\\ \hspace{0.5cm}\(scale=(0.2, 1.0)\)\end{tabular}                                                                              \\ 
\multicolumn{1}{l}{\textbf{Gaussian Blur}}                       & \begin{tabular}[l]{@{}l@{}} \\ \textbf{\textit{For first global view}}\\
\hspace{0.5cm}\(p=1.0\)\\ \textbf{\textit{For second global view}} \\
\hspace{0.5cm}\(p=0.1\)\\ \textbf{\textit{For local views}} \\
\hspace{0.5cm}\(p=0.5\)\end{tabular}            & \begin{tabular}[l]{@{}l@{}}\\ \textbf{\textit{For first global view}}\\
\hspace{0.5cm}\(p=1.0\)\\ \textbf{\textit{For second global view}}\\
\hspace{0.5cm}\(p=0.1\)\\ \textbf{\textit{For local views}} \\
\hspace{0.5cm}\(p=0.5\)\end{tabular}             &
\begin{tabular}[l]{@{}l@{}}\\
\hspace{0.5cm}\(-\)\end{tabular} 
\\ \\
\multicolumn{1}{l}{\textbf{Solarization}}

&\begin{tabular}[l]{@{}l@{}}
\textbf{\textit{For second global view}} \\
\hspace{0.5cm}\(p=0.2\)
\end{tabular}
&
\begin{tabular}[l]{@{}l@{}}
\textbf{\textit{For second global view}}\\
\hspace{0.5cm}\(p=0.2\)  
\end{tabular}
& 
\begin{tabular}[l]{@{}l@{}}\\
\hspace{0.5cm}\(-\)
\end{tabular}       
 \\ 
\multicolumn{1}{l}{\textbf{Horizontal Flip}}  & 
\begin{tabular}[l]{@{}l@{}}
     \hspace{0.5cm}\(p=0.5\) 
\end{tabular} 
 & 
\begin{tabular}[l]{@{}l@{}}
 \hspace{0.5cm}\(p=0.5\) 
\end{tabular}
 & 
\begin{tabular}[l]{@{}l@{}}
     \hspace{0.5cm}\(p=0.5\) 
\end{tabular}

\\ 
\multicolumn{1}{l}{\textbf{Gray Scale}}      & 
\begin{tabular}[l]{@{}l@{}}
    \hspace{0.5cm}\(p=0.2\) 
\end{tabular}
& 
\begin{tabular}[l]{@{}l@{}}
    \hspace{0.5cm}\(p=0.2\) 
\end{tabular}                                                                                                            & 
\begin{tabular}[l]{@{}l@{}}
    \hspace{0.5cm}\(-\) 
\end{tabular}                                                                            \\ 
\multicolumn{1}{l}{\textbf{Color Jittering}}                        &
\begin{tabular}[l]{@{}l@{}}
\hspace{0.5cm}\(p=0.8\)  
\end{tabular}                                                                                                           & 
\begin{tabular}[l]{@{}l@{}}
\hspace{0.5cm}\(p=0.8\)  
\end{tabular}                                                &
\begin{tabular}[l]{@{}l@{}}
\hspace{0.5cm}\(-\)  
\end{tabular}  

\\ \bottomrule
\end{tabular} 
\\ \\
\caption {Data augmentations for pretrained models} \label{tab:data_aug} 
\end{table}

\subsection{Results on more datasets}\label{sec:app-more-datasets}
We have additionally conducted patch classification experiments on ADE20K. As seen in \cref{tab:cs_ade20k_knn_linear}, the ranking of the various ViTs are similar in both k-NN and linear probing settings. 

\Cref{fig:tracking-mot17} shows the results of the tracking experiment performed on MOT17 dataset. 

\begin{figure}
    \centering
    \includegraphics[width=0.6\textwidth]{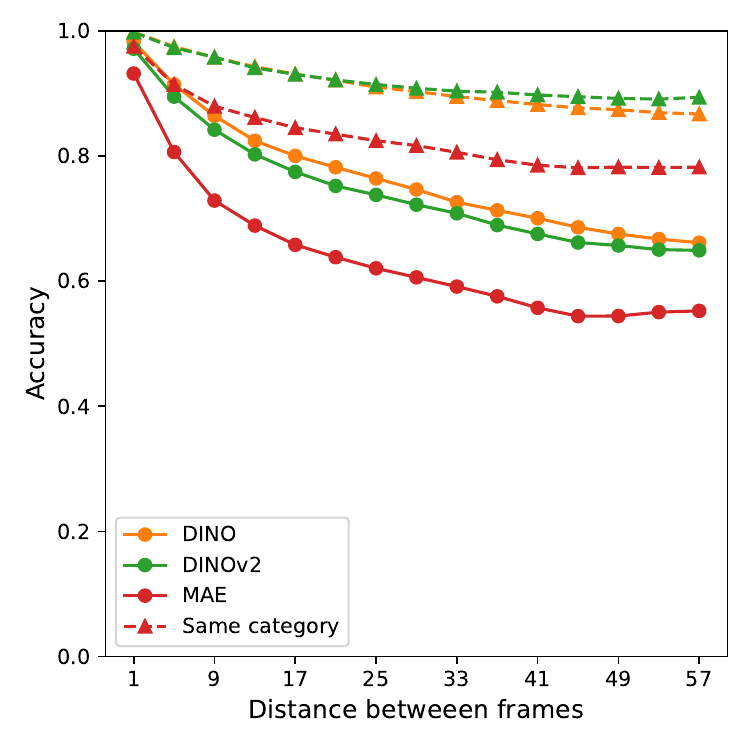}
    \caption{Object instance matching performance for various ViTs on the MOT17 dataset.}
    \label{fig:tracking-mot17}
\end{figure}

\begin{table}[]
\centering
\begin{tabular}{lcccc}
\toprule
    % \multicolumn{2}{c}{cs} & \multicolumn{2}{c}{ade20k}
               & \multicolumn{2}{c}{\textbf{CityScapes}} & \multicolumn{2}{c}{\textbf{ADE20K}}        \\ 
\midrule
               & 1-NN   & Linear & 1-NN    & Linear \\ 
\midrule
\textbf{Supervised ViT} & 0.310 & 0.33   & 0.053  & 0.067  \\
\textbf{DINO}          & 0.345 & 0.383  & 0.055  & 0.070  \\
\textbf{DINOv2}        & 0.496 & 0.590  & 0.136  & 0.161  \\
\textbf{MAE}            & 0.058 & 0.375  & 0.006  & 0.066  \\
\textbf{MAE200}         & 0.295 & -      & 0.043  & 0.056  \\
\textbf{SiMIM}          & 0.075 & 0.156  & 0.006  & 0.032  \\ 
\bottomrule
\end{tabular}
\caption{Patch classification results on CityScapes and ADE20K datasets.}
\label{tab:cs_ade20k_knn_linear}
\end{table}

\subsection{Statistics of the Few-shot Version of FAIR1M Dataset}
We created a subset of the FAIR1M training set in a way that ensures each fine-grained object category appears in at least eight images. We cropped the original images to $224 \times 224$ px tiles and for each tile we kept the list of object categories that are present in the tile. We consider category A to be present in a tile if at least one rotated bounded box of type A has at least 1/3 of its area inside the tile. For each fine-grained category, we took eight images that contain an object of that category. Then we remove those images from the cohort and proceed to the next object category. This way we collected $37 \times 8 - 1$ images, because there were only 7 tiles for one particular fine-grained category (\textit{bus}). \Cref{tab:fair1m-stats} shows the number of patches of each category in our few-shot set of 295 images. Note that DINOv2 has 256 patches per image, while all others have 196 patches per image.

\begin{table}[]
\centering
\vspace{-1.0cm}
{\tiny
\begin{tabular}{ccccc}
\toprule
\textbf{Supercategory}         & \textbf{Fine-grained category} & \textbf{Pixels} & \textbf{Patches} & \textbf{Patches DINOv2} \\ \midrule %\cline{2-5} 
                               & background                     & 13305235        & 51970            & 67897                   \\ \midrule
                               & boeing737                      & 36151           & 139              & 183                     \\ \cmidrule{2-5} 
                               & boeing777                      & 57034           & 228              & 290                     \\ \cmidrule{2-5} 
                               & boeing747                      & 56100           & 218              & 289                     \\ \cmidrule{2-5} 
                               & boeing787                      & 46713           & 178              & 234                     \\ \cmidrule{2-5} 
                               & a320                           & 0               & 0                & 0                       \\ \cmidrule{2-5} 
\multicolumn{1}{c}{}         & a220                           & 42522           & 159              & 220                     \\ \cmidrule{2-5} 
\multicolumn{1}{c}{airplane} & 576                            & 787             & 269              & 350                     \\ \cmidrule{2-5} 
                               & a350                           & 100574          & 398              & 506                     \\ \cmidrule{2-5} 
                               & a321                           & 57768           & 228              & 293                     \\ \cmidrule{2-5} 
                               & c919                           & 14654           & 57               & 74                      \\ \cmidrule{2-5} 
                               & arj21                          & 11891           & 51               & 67                      \\ \cmidrule{2-5} 
                               & other-airplane                 & 62174           & 246              & 312                     \\ \cmidrule{2-5} 
                               & total                          & 554797          & 2171             & 2818                    \\ \midrule
                               & passenger ship                 & 26988           & 101              & 134                     \\ \cmidrule{2-5} 
                               & motorboat                      & 28031           & 118              & 142                     \\ \cmidrule{2-5} 
                               & fishing boat                   & 30738           & 127              & 160                     \\ \cmidrule{2-5} 
                               & tugboat                        & 10226           & 42               & 55                      \\ \cmidrule{2-5} 
\multicolumn{1}{c}{ship}     & engineering ship               & 88900           & 342              & 443                     \\ \cmidrule{2-5} 
                               & liquid cargo ship              & 24148           & 91               & 122                     \\ \cmidrule{2-5} 
                               & dry cargo ship                 & 144103          & 562              & 732                     \\ \cmidrule{2-5} 
                               & warship                        & 28255           & 114              & 142                     \\ \cmidrule{2-5} 
                               & other-ship                     & 15820           & 63               & 81                      \\ \cmidrule{2-5} 
                               & total                          & 397209          & 1560             & 2011                    \\ \midrule 
                               & small car                      & 24371           & 98               & 125                     \\ \cmidrule{2-5} 
                               & bus                            & 2451            & 8                & 13                      \\ \cmidrule{2-5} 
                               & cargo truck                    & 22516           & 92               & 119                     \\ \cmidrule{2-5} 
                               & dump truck                     & 8553            & 37               & 44                      \\ \cmidrule{2-5} 
\multicolumn{1}{c}{}         & van                            & 33960           & 131              & 165                     \\ \cmidrule{2-5} 
\multicolumn{1}{c}{car}      & trailer                        & 7654            & 32               & 39                      \\ \cmidrule{2-5} 
                               & tractor                        & 2734            & 10               & 16                      \\ \cmidrule{2-5} 
                               & truck tractor                  & 3760            & 17               & 17                      \\ \cmidrule{2-5} 
                               & excavator                      & 10048           & 38               & 52                      \\ \cmidrule{2-5} 
                               & other-vehicle                  & 8138            & 37               & 47                      \\ \cmidrule{2-5} 
                               & total                          & 124185          & 500              & 637                     \\ \midrule 
                               & baseball field                 & 108367          & 421              & 556                     \\ \cmidrule{2-5} 
\multicolumn{1}{c}{}         & basketball court               & 16005           & 72               & 82                      \\ \cmidrule{2-5} 
\multicolumn{1}{c}{court}    & football field                 & 87005           & 338              & 438                     \\ \cmidrule{2-5} 
                               & tennis court                   & 57114           & 212              & 294                     \\ \cmidrule{2-5} 
                               & total                          & 268491          & 1043             & 1370                    \\ \midrule 
                               & roundabout                     & 94068           & 360              & 490                     \\ \cmidrule{2-5} 
\multicolumn{1}{c}{road}     & intersection                   & 38180           & 144              & 190                     \\ \cmidrule{2-5} 
                               & bridge                         & 19755           & 72               & 107                     \\ \cmidrule{2-5} 
                               & total                          & 152003          & 576              & 787                     \\ 
                               \bottomrule

\end{tabular}
}

\vspace{0.3cm}

\caption{Patch statistics of the few-shot FAIR1M subset we used in our experiments.}
\label{tab:fair1m-stats}
\end{table}

\subsection{Reconstruction Error Analysis for MAE}
To understand what information is stored in the high variance features of MAE, if removing them does not harm patch classification or patch retrieval performance, we conduct experiments with image reconstruction. The hypothesis suggests that the removed features play a role in certain reconstruction properties. We use the pretrained decoder of MAE in two settings: when no patches are masked and when 75\% of the patches are masked. In~\cref{tab:mae-reconstruct}, one can see that when the high variance features are filled with zeros, the reconstruction metrics get slightly worse. This indicates that these features contain knowledge on how to reconstruct the image, but they are not essential for most other downstream tasks. The accuracy of the reconstruction is evaluated using Mean Square Error (MSE), Peak Signal-to-Noise Ratio (PSNR) and Structural Similarity Index (SSIM)~\cite{ssim} metrics.

\begin{table}[]
\centering
\begin{tabular}{llccc}
\toprule
                        &                   & \begin{tabular}[c]{@{}c@{}}Original\\ Reconstruction\end{tabular} & \begin{tabular}[c]{@{}c@{}}Without high\\ variance features\end{tabular} & \begin{tabular}[c]{@{}c@{}}Without low\\ variance features\end{tabular} \\
\midrule
\multicolumn{1}{c}{MSN $\downarrow$} & No masking        & 0.163                                                             & 0.199                                                                    & 1.173                                                                   \\
                        & Mask ratio = 75\% & 0.090                                                             & 0.103                                                                    & 1.164                                                                   \\
\midrule
PSNR  $\uparrow$                  & No masking        & 25.666                                                            & 23.349                                                                   & 15.350                                                                  \\
                        & Mask ratio = 75\% & 25.571                                                            & 24.488                                                                   & 14.474                                                                  \\
\midrule
SSIM  $\uparrow$                  & No masking        & 0.78                                                              & 0.722                                                                    & -0.074                                                                  \\
                        & Mask ratio = 75\% & 0.786                                                             & 0.737                                                                    & -0.064          \\
\bottomrule
\end{tabular}
\vspace{0.3cm}
\caption{Reconstruction metrics for MAE with and without replacement of 200 highest variance and 200 lowest variance features with zeros.}
\label{tab:mae-reconstruct}
\end{table}

\subsection{Frequency Noise Computation}

In \cref{fig:segmentation-degradation} we presented some degradation analyses and results for various degradations, including frequency-based random noise. In this section, we briefly expand on the experiments and go over their setup. 

To create frequency-based random noise, we first generated 2D Gaussian random noise with the same dimensions as the image across all three color channels. We then applied a Fourier transform to the noise, masked it in the frequency space, and applied an inverse Fourier transform to obtain the frequency-based random noise. The following formula demonstrates the addition of frequency-based random noise to the image:

$$
I = I_{0} + \mathcal{F}^{-1}(\mathcal{F}(\epsilon) \odot \mathbf{M}_f),
$$
where $I_{0}$ corresponds to the original image, $\epsilon$ corresponds to random noise with the same dimensions as the image. 
Each pixel of the noise follows a Gaussian distribution with a mean of $0$ and a given variance, $M$ represents the frequency mask shown in \cref{fig:frequency_based_random_noise}, we also shifted the zero-frequency component to the center of the spectrum and then applied an inverse shift. At last, $\mathcal{F}$ and $\mathcal{F}^{-1}$ correspond to the Fourier transform and inverse Fourier transform, respectively.
The additive frequency-based random noise, the corresponding mask, and the noisy images for four different makes are demonstrated in~\cref{fig:frequency_based_random_noise} for more details.

\begin{figure}
    \centering
    \includegraphics[width=\linewidth]{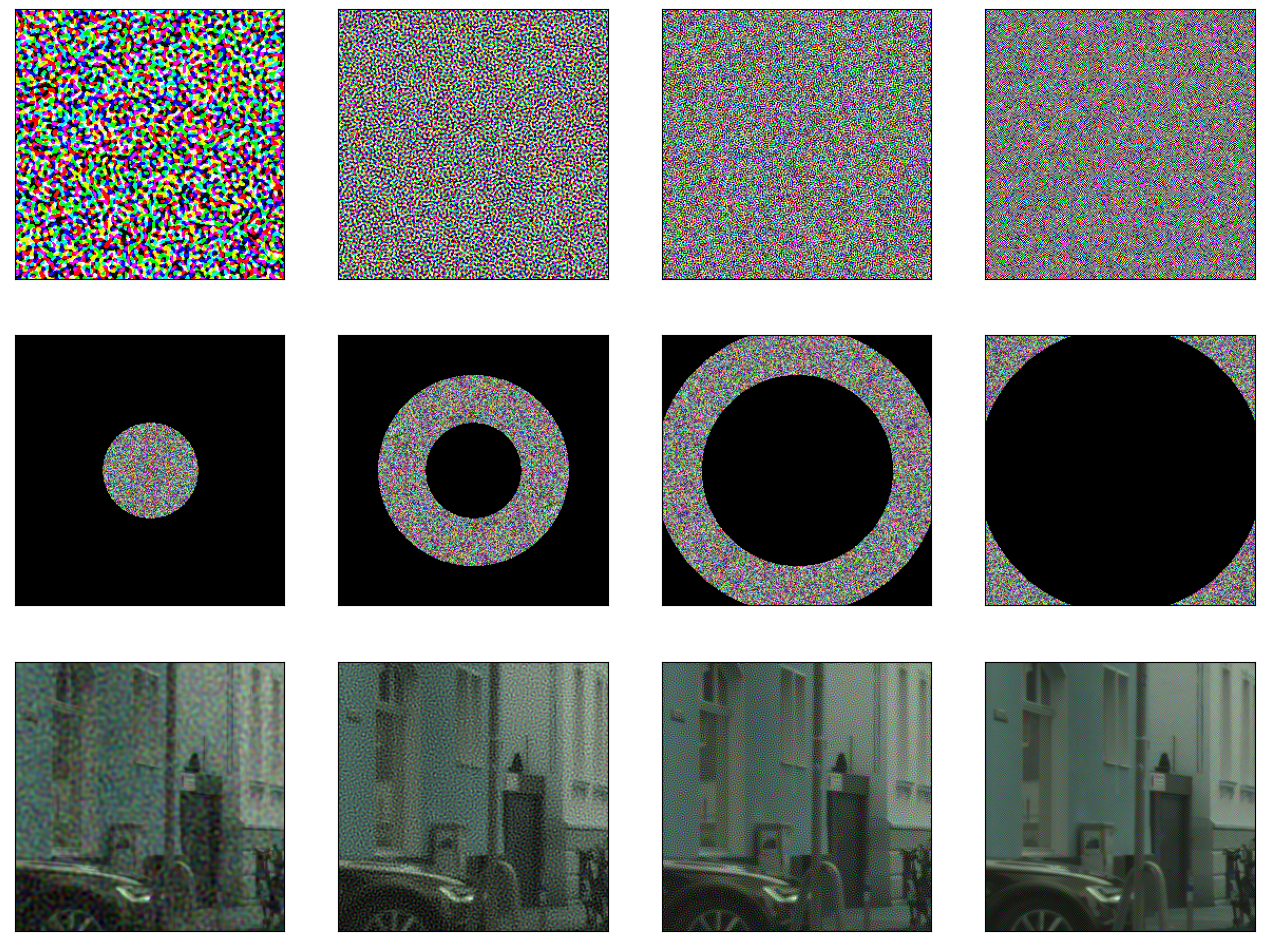}
    \caption{The first row shows the additive noise, the second row corresponds to the masked noise in the frequency domain, and the last row corresponds to the image with additive frequency-based noise.}
    \label{fig:frequency_based_random_noise}
\end{figure}

\subsection{Additional Experiments on FAIR1M}
\subsubsection{DINO vs. iBOT vs. DINOv2}
The most unexpected result of \cref{sec:results} is that DINO representations are better for retrieving the closest patch given a corrupted patch than DINOv2 representations. DINOv2 has a series of differences compared to DINO. These differences can be split into two categories: those related to the loss terms and those related to the scale of the model and the dataset. The new patch-level loss term of DINOv2 first appeared in iBOT. Here we performed the same set of experiments on iBOT as well, to compare it with DINO and DINOv2. As seen in \cref{fig:fair1m-ibot}, iBOT performs at least as good as DINO. This means that the new loss term cannot be blamed for the worse retrieval performance for DINOv2.

\begin{figure}
    \centering
    \includegraphics[width=0.8\linewidth]{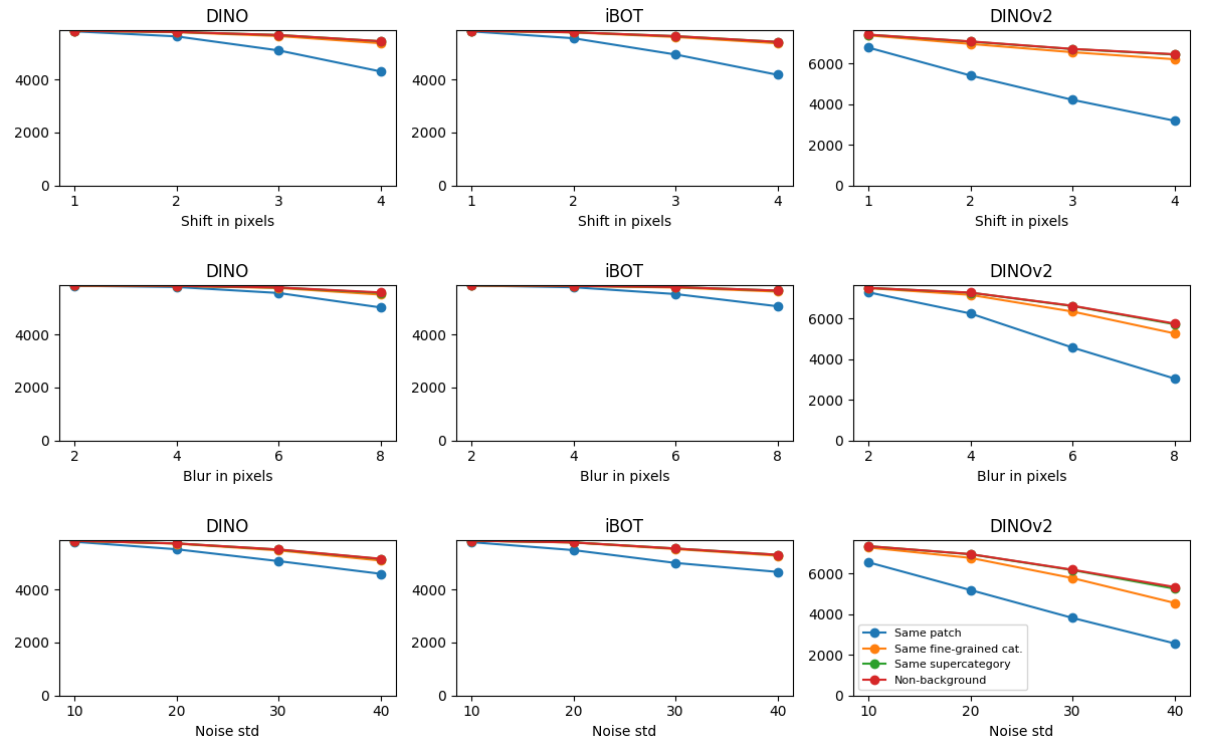}
    \caption{iBOT's retrieval performance is slightly better than DINO, and is much better than DINOv2.}
    \label{fig:fair1m-ibot}
\end{figure}

\subsection{Discussion on Tiling}

Since most ViTs are trained on small-sized images and their original weights are provided for $224 \times 224$ images without interpolating positional embeddings, preprocessing becomes necessary when working with larger images. For example, in the case of the FAIR1M dataset, the images are $1000 \times 1000$, and for CityScapes the images are $1024 \times 2018$. There are several options for handling such images, including rescaling them to a smaller size, tiling them (dividing them into smaller pieces and conducting experiments on these tiles, then combining them to reconstruct the original image size), or forcing the ViT to process the full-sized image by interpolating the positional embeddings.
The risk of tiling the images is that smaller tiles might lose the global context of the image, which is an important factor in transformer architecture. In all experiments in the paper we chose the tiling method. Here we explore the effect of using full-scale images.

\begin{table}[]
\centering

\begin{tabular}{lccc}
\toprule

                                & \textbf{Input Size}  & \textbf{Linear} & \textbf{KNN}  \\
\midrule
\textbf{DINO} & $256\times256$  & \textbf{0.383}  & \textbf{0.345} \\ 
\textbf{DINO} & $1024\times2048$ & 0.340  & 0.308 \\
\textbf{MAE} & $256\times256$   & \textbf{0.375}  & \textbf{0.058} \\
\textbf{MAE} & $1024\times2048$ & 0.176  & 0.02 \\
\bottomrule
\end{tabular}
\vspace{0.3cm}
\caption{Patch classification performance on Cityscapes depending on tiling of the input images.}
\label{tab:full_vs_tiled}

\end{table}

\begin{table}[]
\centering
\begin{tabular}{lcccc}
\toprule
% \hline
           & \textbf{MAE}    & \textbf{MAE200} & \textbf{MAE Random} & \textbf{MAE Junk} \\ 
\midrule
road       & 0.606  & 0.603  & 0.588      & 0.551    \\ 
building   & 0.698  & 0.711  & 0.678      & 0.509    \\ 
vegetation & 0.723  & 0.700  & 0.687      & 0.591    \\ 
\midrule
mean       & 0.6754 & 0.671  & 0.651      & 0.550    \\ 
\bottomrule
\end{tabular}
\vspace{0.3cm}
\caption{$R^2$ for MAE for different classes for global context understanding.}
\label{tab:r_squared}
\end{table}

\subsubsection{Cityscapes}

We conduct the following two experiments. For the first experiment, we tile the images into $256\times256$ patches, resize them into $224 \times 224$, and separately compute the embeddings for corresponding ViTs (only for DINO and MAE).
For the second experiment, we compute the patch embeddings for the full images.
We check the accuracy of semantic segmentation for CityScapes dataset on a validation set of 30 images.
Surprisingly, both MAE and DINO for linear probing and k-NN achieve higher mIoU values in the tiled setting. The results are summarized in~\cref{tab:full_vs_tiled}
In conclusion, despite the expectation that the global context of the image would contain more information, the degradation of the performance due to large input sizes is too strong.

\subsubsection{Tracking}
We perform a similar experiment for the object tracking setup. Note that in this setting we pool the representations of all patches inside a bounding box. If the object is split into multiple tiles of the same image, the averaging will occur over patch embeddings from different tiles. In \cref{fig:tracking-full-vs-tiled} we see the above phenomenon for DINO and DINOv2, tiled images perform better. For MAE we see a surprising result, object representations pulled from embeddings of the full image perform better.

\begin{figure}
    \centering
    \includegraphics[width=0.6\textwidth]{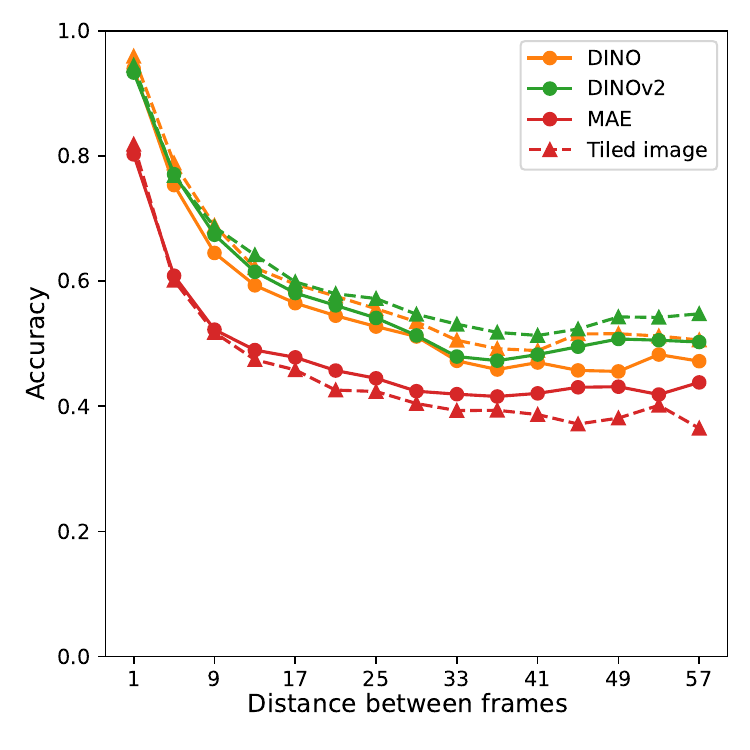}
    \caption{Comparison of full image and tiled versions of MAE and DINO in the object tracking experiment}
    \label{fig:tracking-full-vs-tiled}
\end{figure}

Note that all experimental results in the object tracking experiments in this paper are reported on a subset of 4 videos from BDD-100k.

\subsection{Sample Predictions on Cityscapes}

In \cref{fig:cs-knn-linear}, we present qualitative results of how semantic segmentation looks based on the ViTs used in this work. We fixed two images from the Cityscapes dataset. See the first and third figures of the first row in~\cref{fig:cs-knn-linear} for the examples and their corresponding original masks, which can be seen in the second and fourth figures of the first row, respectively. The second, third, and subsequent rows of~\cref{fig:cs-knn-linear} demonstrate the segmentation masks obtained by the corresponding ViT, where the first and third columns correspond to the k-NN-based prediction and the second and fourth columns correspond to the linear probing-based prediction for the corresponding instances from the Cityscapes dataset.
In these figures, we can qualitatively reconfirm our observations that MAE almost completely fails to segment the patches correctly with k-NN. However, its performance is competitive with linear probing. We also observe that MAE-200, which corresponds to the embeddings obtained by MAE without the top 200 features of the largest variance, outperforms MAE for k-NN and is almost identical to that of MAE for linear probing. As expected, DINO and DINOv2 qualitatively outperform all the other methods.

\begin{figure}
    \centering
    \includegraphics[width=\linewidth]{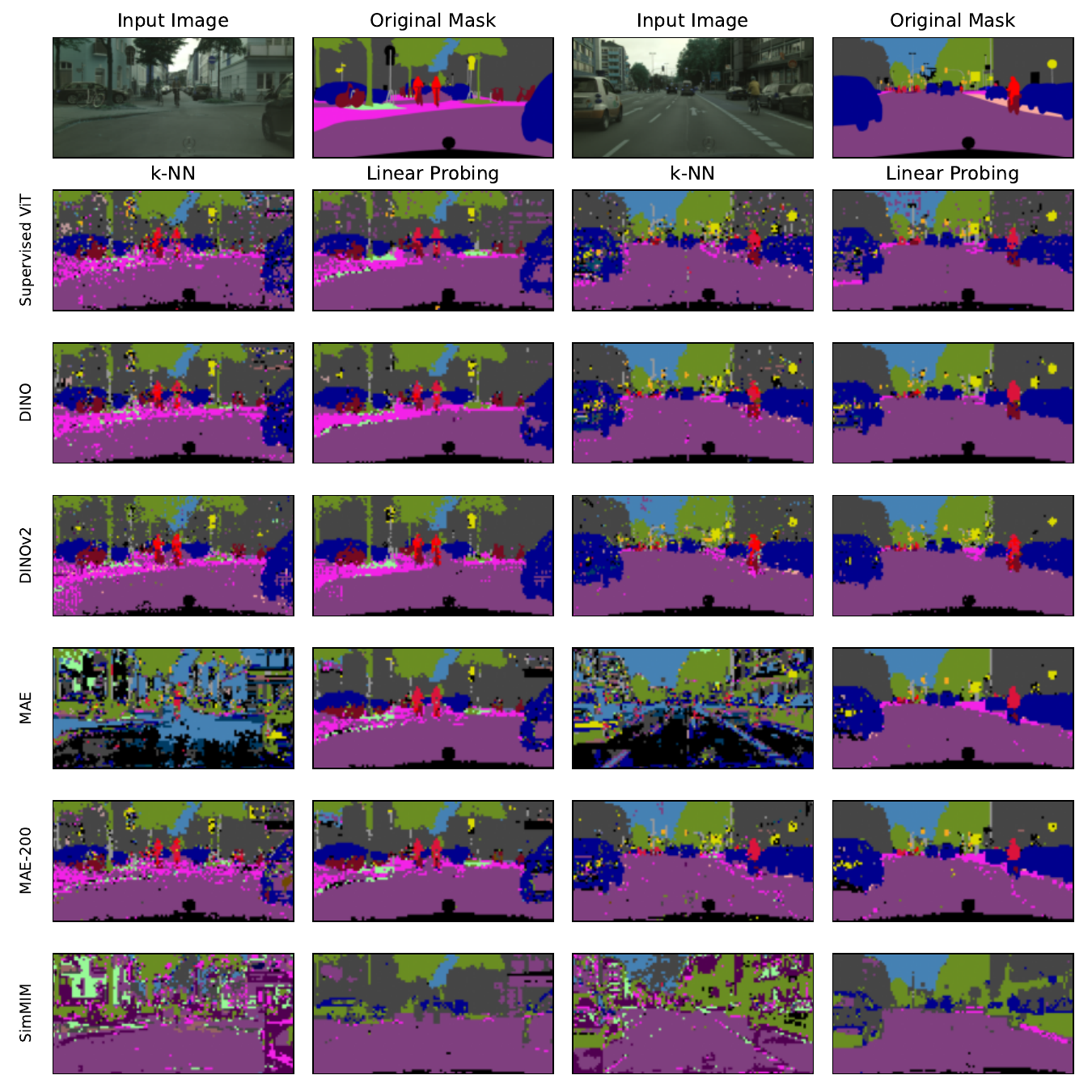}
    \caption{Predictions of k-NN and linear probing for all models for two selected images from Cityscapes.}
    \label{fig:cs-knn-linear}
\end{figure}

 \fi

\end{document}